\documentclass[Afour,sageh,times]{sagej}

\usepackage{moreverb,url}
\usepackage{algorithm, setspace}
\usepackage{algpseudocode}
\usepackage{csquotes}
\usepackage{array,multirow}
\usepackage{tabularx}
\usepackage{adjustbox}
\usepackage{multicol}
\usepackage{subfigure}
\usepackage{color, colortbl}
\usepackage{xcolor}

\definecolor{Gray}{gray}{0.9}

\newcommand\BibTeX{{\rmfamily B\kern-.05em \textsc{i\kern-.025em b}\kern-.08em
T\kern-.1667em\lower.7ex\hbox{E}\kern-.125emX}}

\newcommand{\adnote}[1]%
 {\textcolor{blue}{\textbf{AD: #1}}}
 \newcommand{\abnote}[1]%
 {\textcolor{red}{\textbf{AB: #1}}}
 \newcommand{\dlnote}[1]%
 {\textcolor{green}{\textbf{DL: #1}}}
 
\usepackage{hyperref}
 \hypersetup{
     colorlinks=true,
     linkcolor=orange,
     filecolor=orange,
     citecolor=orange,      
     urlcolor=orange,
     }
\usepackage{perpage} 
\MakePerPage{footnote} 

\begin{document}

\runninghead{Losey et al.}

\title{Physical Interaction as Communication: \\ Learning Robot Objectives Online from Human Corrections}

\author{Dylan P. Losey\affilnum{1}, Andrea Bajcsy\affilnum{2}, Marcia K. O'Malley\affilnum{3}, and Anca D. Dragan\affilnum{2}}

\affiliation{\affilnum{1}Virginia Tech; losey@vt.edu\\
\affilnum{2}University of California, Berkeley; \{abajcsy, anca\}@berkeley.edu\\\affilnum{3}Rice University; omalleym@rice.edu}

\corrauth{Dylan P. Losey, Department of Mechanical Engineering, Virginia Tech, 635 Prices Fork Road, Blacksburg, VA 24061}

\email{losey@vt.edu}

\begin{abstract}


When a robot performs a task next to a human, physical interaction is inevitable: the human might push, pull, twist, or guide the robot. The state-of-the-art treats these interactions as \emph{disturbances} that the robot should reject or avoid. At best, these robots respond safely while the human interacts; but after the human lets go, these robots simply return to their original behavior. We recognize that physical human-robot interaction (pHRI) is often intentional---the human intervenes on purpose because the robot is not doing the task correctly. In this paper, we argue that when pHRI is \emph{intentional} it is also \emph{informative}: the robot can leverage interactions to learn how it should complete the rest of its current task even after the person lets go. We formalize pHRI as a dynamical system, where the human has in mind an objective function they want the robot to optimize, but the robot does not get direct access to the parameters of this objective---they are internal to the human. Within our proposed framework human interactions become observations about the true objective. We introduce approximations to \emph{learn from} and \emph{respond to} pHRI in real-time. We recognize that not all human corrections are perfect: often users interact with the robot noisily, and so we improve the efficiency of robot learning from pHRI by reducing unintended learning. Finally, we conduct simulations and user studies on a robotic manipulator to compare our proposed approach to the state-of-the-art. Our results indicate that learning from pHRI leads to better task performance and improved human satisfaction.

\end{abstract}


\maketitle

\section{Introduction}

Physical interaction is a natural means for collaboration and communication between humans and robots. From compliant designs to reliable prediction algorithms, recent advances in robotics have enabled humans and robots to work in close physical proximity. Despite this progress, seamless physical interaction---where robots are as responsive, intelligent, and fluid as their human counterparts---remains an open problem. 

One key challenge is determining how robots should \textit{respond} to direct physical contact. Fast and safe responses to external forces are generally necessary, and have been studied extensively within the field of physical human-robot interaction (pHRI). A traditional controls approach is to treat the human's interaction force as a perturbation to be rejected or ignored. Here the robot assumes that it is an expert agent and follows its own predefined trajectory regardless of the human's actions \citep{de2008atlas}. Alternatively, the robot can treat the human as the expert, so that the human guides the passive robot throughout their preferred trajectory. Whenever the robot detects an interaction it stops moving and becomes transparent, enabling the human to easily adjust the robot's state \citep{jarrasse2012framework}. Impedance control---the most prevalent paradigm for pHRI \citep{haddadin2016physical, hogan1985impedance}---combines aspects of the previous two control strategies. Here the robot tracks a predefined trajectory, but when the human interacts the robot complies with the human's applied force. Under this approach the human can intuitively alter the robot's state while also receiving force feedback from the robot.

In each of these different response strategies for pHRI the robot returns to its pre-planned trajectory as soon as the human stops interacting. In other words, the robot remains confident that its original trajectory is the correct way to complete the task. Since this robot trajectory is optimal with respect to some underlying objective function, these response paradigms effectively maintain a fixed objective function during pHRI. Hence, the human's interactions do not change the robot's understanding of the task; instead, external forces are simply \emph{disturbances} which should be reacted to, rather than information which should be reasoned about.

In this work we assert that physical human interactions are often \textit{intentional}, and occur because the robot is doing something that the human believes is incorrect. The fact that the human is physically intervening to fix the robot's behavior implies that the robot's trajectory---and therefore the underlying objective function used to produce this trajectory---is wrong. Under our framework we consider the forces that the human applies as observations about the true objective function that the robot should be optimizing, which is known to the human but not by the robot. Accordingly, human interactions should no longer be thought of as only disturbances that perturb the robot from its pre-planned trajectory, but rather as \textit{corrections} that teach the robot about the desired behavior during the task.





This insight enables us to formalize the robot's response to pHRI as an instance of a partially observable dynamical system, where the robot is unsure of its true objective function, and human interactions provide information about that objective. Solving this system defines the \emph{optimal way} for the robot to respond to pHRI. We derive an approximation of the solution to this system that works in real-time for continuous state and action spaces, enabling robot arms to react to pHRI online and adjust how they complete the current task. Due to the necessity of fast and reactive schemes, we also derive an online gradient-descent solution that adapts inverse reinforcement learning approaches to the pHRI domain. We find that this solution works well in some settings, while in others user corrections are noisy and result in unintended learning. We alleviate this problem by introducing a restriction to our update rule focused on extracting only what the person intends to correct, rather than assuming that every aspect of their correction is intentional. Finally, we compare our approximations to a full solution, and experimentally test our proposed learning method in user studies with a robotic manipulator.


We make the following contributions\footnote{Note that parts of this work have been published at the Conference on Robotic Learning \citep{bajcsy2017online} and the Conference on Human-Robot Interaction \citep{bajcsy2018learning}.}:

\smallskip


\noindent\textbf{Formalizing pHRI as implicitly communicating objectives.} We formalize reacting to physical human-robot interaction as a \emph{dynamical system}, where the robot optimizes an objective function with an unknown parameter $\theta$, and human interventions serve as observations about the true value of $\theta$. As posed, this problem is an instance of a Partially Observable Markov Decision Process (POMDP). 
    
\smallskip


\noindent\textbf{Learning online from pHRI and safely controlling the robot.} Responding to pHRI requires learning about the objective in real-time (the estimation problem), as well as adapting the robot's motion in real-time (the control problem). We derive an approximation that enables both by moving from the \emph{action or policy} level to the \emph{trajectory} level, bypassing the need for dynamic programming or POMDP solvers, and instead relying on local optimization. Working at the trajectory level we derive an online gradient descent learning rule which updates the robot's estimate of the true objective $\theta$ as a function of the human's interaction force.



\begin{figure}[!t]
\centering
\subfigure[Robot that treats physical interactions as disturbances.]{\includegraphics[width=.5\textwidth]{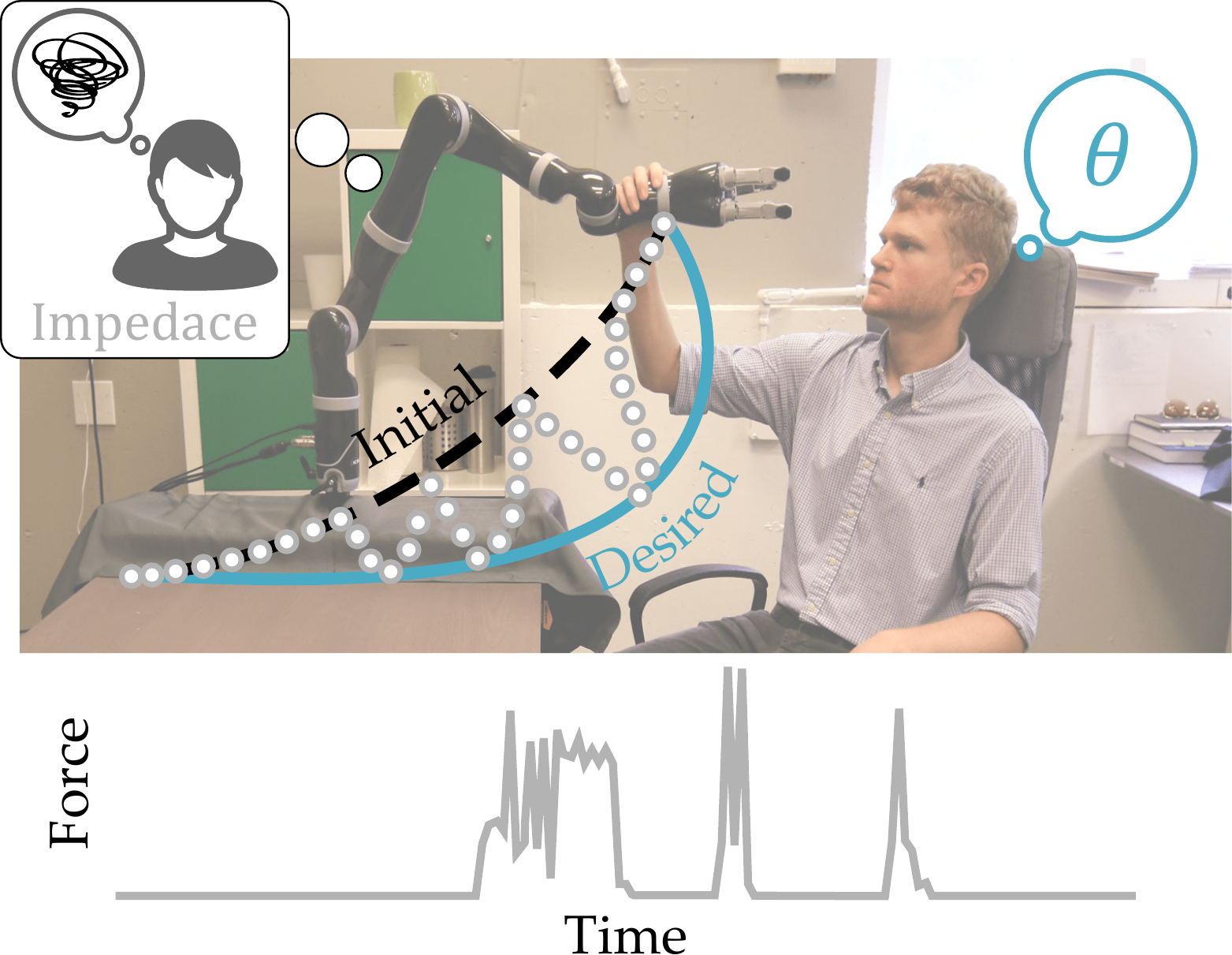}}

\subfigure[Robot that treats physical interactions as intentional and informative.]{\includegraphics[width=.5\textwidth]{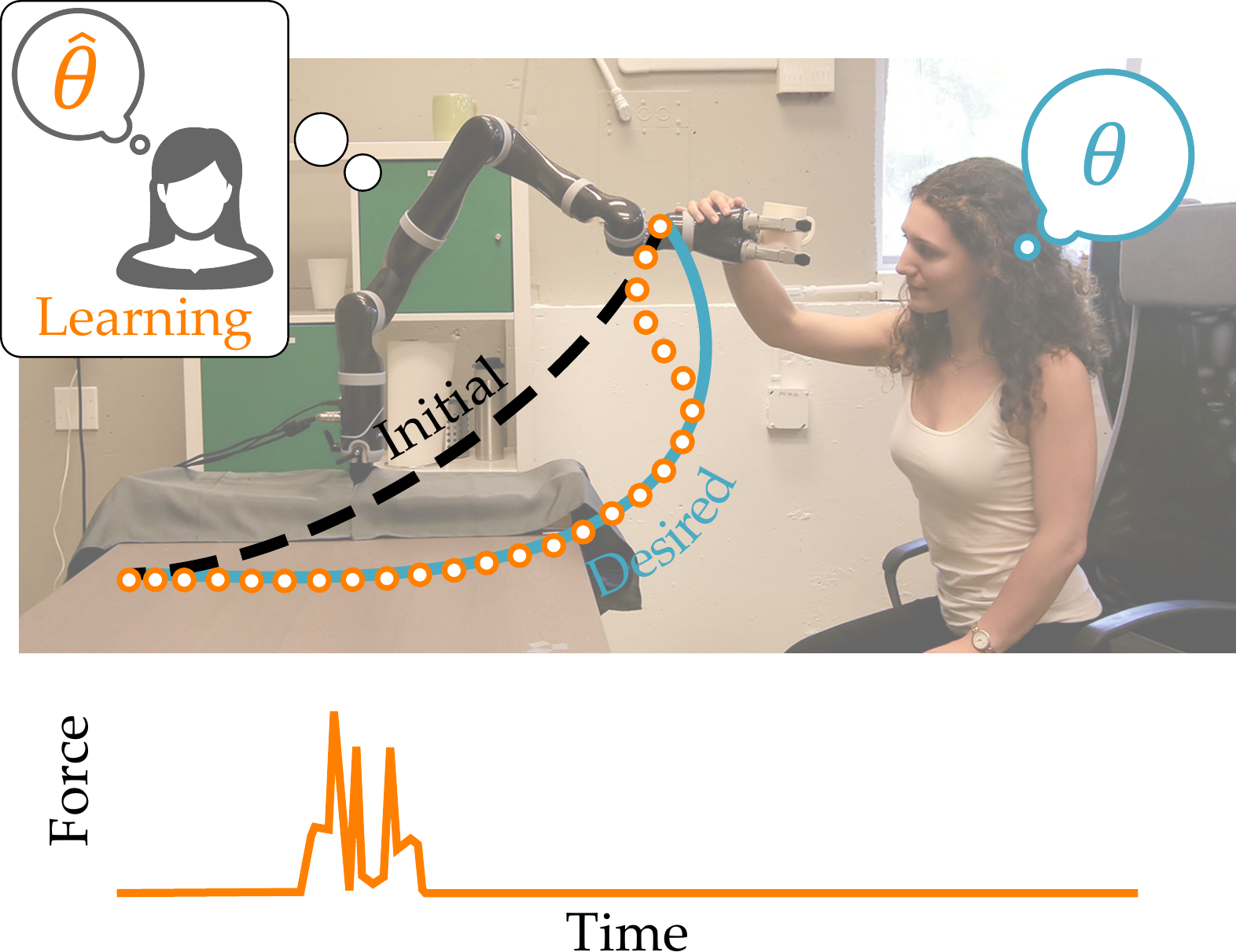}}

\caption{(Top) When physical human interactions are treated as disturbances people have to repeatedly push the robot to physically change its behavior. (Bottom) Robots that recognize that physical interactions may be corrections can learn from these interactions and change their underlying behavior to align with the human's preferences.}
\label{fig:intro}
\end{figure}

\smallskip

\noindent\textbf{Responding to unintended human corrections.} In practice, the human's physical interactions are noisy and imperfect, particularly when trying to correct high degree-of-freedom (DoF) robotic arms. Because these corrections do not isolate exactly what the human is trying to change, responding to all aspects of pHRI can result in \emph{unintended learning}. We therefore introduce a restriction to our online learning rule that only updates the robot's estimate over aspects of the task that the person was most likely trying to correct.

\smallskip


\noindent\textbf{Analyzing approximate solutions.} In a series of controlled human-robot simulations we compare the performance of our online learning algorithm to the gold standard: computing an optimal \emph{offline} solution to the pHRI formalism. We also consider two baselines: \emph{deforming} the robot's original trajectory in the direction of human forces, and reacting to human forces with only \emph{impedance control}. We find that our online learning method outperforms the deformation and impedance control baselines, and that the difference in performance between our online learning method and the more complete offline solution is negligible.   

\smallskip


\noindent\textbf{Conducting user studies on a 7-DoF robot.} We conduct two user studies with the JACO2 (Kinova) robotic arm to assess how online learning from physical interactions affects the robot's objective performance and the user's subjective feedback. During these studies the robot begins with an incorrect objective function and participants must physically intervene mid-task to teach the robot to execute the remainder of the task correctly. In our first study we find that participants are able to physically teach the to perform the task correctly, and that participants prefer robots that learn from pHRI. In our second study we test how learning from all aspects of the human's interaction compares to our restriction, where the robot only learns about the single feature most correlated with the human's correction.

\smallskip



Overall, this work demonstrates how we can leverage the implicit communication which is present during physical interactions. Learning from implicit human communication applies not only to pHRI, but conceivably also to other kinds of actions that people take.

\section{Prior Work}



In this work, we enable robots to leverage physical interaction with a human during task execution to learn a human's objective function. We also account for imperfections in the way that people physically interact to correct robot behavior. Prior work has separately addressed (a) control strategies for reacting to pHRI \emph{without} learning the human's objective and (b) learning the human's objective \emph{offline} from kinesthetic demonstrations. An exception is work on shared autonomy, which learns the human's objective in real-time, but only when that objective is parameterized by the human's goal position. Finally, we discuss related work on algorithmic teaching, which describes how humans can optimally teach robots as well as how humans practically teach robots. \medskip


\noindent \textbf{Controllers for pHRI}. Recent review articles on control for physical human-robot interaction \citep{haddadin2016physical,de2008atlas} group these controllers into three categories: impedance control, reactive strategies, and shared control. When selecting a controller for pHRI, ensuring the human's safety is crucial. Impedance control, as originally proposed by \cite{hogan1985impedance}, achieves human safety by making robots compliant during interactions; for instance, the robot behaves like a spring-damper centered at the desired trajectory. But the robot can react to human contacts in other ways besides---or in addition to---rendering a desired impedance. \cite{haddadin2008collision} suggest a variety of alternatives: the robot could stop moving, switch to a low-impedance mode, move in the direction of the human's applied force, or re-time its desired trajectory.

 
More relevant here are works on shared control, where the robot has an objective function, and uses that objective function to select optimal control feedback during pHRI \citep{jarrasse2012framework, medina2015synthesizing,losey2018review}. In \cite{li2016framework} the authors formulate pHRI with game theory. The robot has an objective function which depends on the error from a pre-defined trajectory, the human's effort, and the robot's effort. During the task the robot learns the relative weights of these terms from human interactions, resulting in a shared controller that becomes less stiff when the human exerts more force. Rather than only learning the correct robot stiffness---as in \cite{li2016framework}---our work more generally learns the correct robot behavior. We note that each of these control methods \citep{hogan1985impedance,haddadin2008collision,jarrasse2012framework,medina2015synthesizing,li2016framework,losey2018review} enables the robot to safely respond to human interactions in real-time. However, once the human stops interacting, the robot resumes performing its task \emph{in the same way} as it had planned before human interactions. \medskip


\noindent \textbf{Learning Human Objectives Offline}. Inverse reinforcement learning (IRL), also known as inverse optimal control, explicitly learns the human's objective function from demonstrations \citep{abbeel2004apprenticeship,kalman1964linear,ng2000algorithms,osa2018algorithmic}. IRL is an instance of supervised learning where the human shows the robot the correct way to perform the task, and the robot infers the human's objective \emph{offline} from one or more demonstrations. Demonstrations can be provided through pHRI, where the human kinesthetically guides the passive robot along their desired trajectory \citep{finn2016guided,kalakrishnan2013learning}. In practice, the human's actual demonstrations may not be optimal with respect to their objective, and \cite{ramachandran2007bayesian,ziebart2008maximum} address IRL from approximately optimal or noisy demonstrations. 

Most relevant to our research are IRL approaches that learn from corrections to the robot's trajectory rather than complete demonstrations \citep{jain2015learning,karlsson2017demonstrations,ratliff2006maximum}. Within these works, the human corrects some aspect of the demonstrated trajectory during the current iteration, and the robot improves its trajectory \emph{the next time} it performs the task. By contrast, we use human interactions to update the robot's behavior during the \emph{current} task. Our solution for real-time learning is analogous to online Maximum Margin Planning \citep{ratliff2006maximum} or coactive learning \citep{jain2015learning,shivaswamy2015}, but we derive this solution as an approximately optimal response to pHRI. Moreover, we also show how this learning method can be adjusted to accommodate \emph{unintentional} human corrections.

\textcolor{black}{As we move towards online learning, we also point out research where the robot learns a discrete set of candidate reward functions offline, and then changes between these options based on the human’s real-time physical corrections \citep{yin2019ensemble}. We view this work as a simplified instance of our approach, where the robot has sufficient domain knowledge to limit the continuous space of rewards to a few discrete choices.}

\medskip

\noindent \textbf{Learning Human Goals Online}. Prior work on shared autonomy has explored how robots can learn the human's objective \emph{online} from the human's actions. \cite{dragan2013policy,javdani2018shared} consider human-robot collaboration and teleoperation applications, in which the robot observes the human's inputs, and then infers the human's desired \emph{goal position} during the current task. Other works on shared autonomy have extended this framework to learn the human's adaptability \citep{nikolaidis2017human} or trust \citep{chen2018planning} so that the robot can reason about how its actions may alter the human's goal. In all of these prior works the robot is moving through free-space and the human's preferred goal is the only aspect of the true objective which is unknown. We build on this prior work by considering \emph{general objective parameters}; this requires a more complex---i.e., non-analytic and difficult to compute---observation model, along with additional approximations to achieve online performance. 

\textcolor{black}{Although not part of shared autonomy, we also point out research where the robot's trajectory changes online due to physical human interactions. In some works---such as \cite{mainprice2013human,sisbot2007human}---the robot alters its trajectory to \textit{avoid} physical human interaction. More related to our approach are works where the robot \textit{embraces} physical corrections to adapt its behavior. For example, in \cite{losey2018trajectory, khoramshahi2018human, khoramshahi2019dynamical,losey2019learning} the robot maintains a parameterized desired trajectory or dynamical system, and updates the parameters in real-time to minimize the error between the resultant trajectory and the human's corrections. These works directly update the robot’s desired trajectory based on corrections; by contrast, we learn a \textit{reward function} from human corrections, which can---in turn---be used to generate dynamical systems or desired trajectories. Learning a reward function is advantageous here because it enables the robot to generalize what it has learned within the task, e.g., because the human has corrected the robot closer to one table, the robot will move closer to a second table as well.}

\medskip

\noindent \textbf{Humans Teaching Robots}. Recent works on algorithmic teaching, also referred to as machine teaching, can be used to find the optimal way to teach a learning agent \citep{balbach2009recent,goldman1995complexity,zhu2015machine}. Within our setting the human teaches the robot their objective function via corrections, but actual end-users are imperfect teachers. Algorithmic teaching addresses this issue by improving the human's demonstrations for IRL \citep{cakmak2012algorithmic}. Here the robot learner provides advice to the human teacher, guiding them into making better corrections. By contrast, we focus on developing learning algorithms that match how \emph{everyday end-users} approach the task of teaching \citep{thomaz2008teachable,thomaz2009learning,jonnavittula2020know}. Put another way, we do not want to optimize the human's corrections, but rather develop learning algorithms that account for imperfect teachers. Most relevant is \cite{akgun2012keyframe}, which shows how humans can kinesthetically correct the robot's waypoints offline to better match their desired trajectory. We similarly investigate interfaces that make it easier for people to teach robots, but in the context of applying physical forces to correct an existing robot trajectory.
\section{Formalizing Physical Human-Robot Interaction}\label{sec:formalism}

Consider a robot performing a task autonomously and in close proximity to a human end-user. The human observes this robot and can \emph{physically} interact with the robot to alter its behavior. Returning to our running example from Fig.~\ref{fig:intro}, imagine a robotic manipulator that is carrying a coffee mug from the top of a cabinet down to a table while the human sits nearby. Importantly, the robot is either not doing this task correctly (e.g., the robot is carrying the cup at such an angle that coffee will spill) or the robot is not doing the task according to the human's personal preferences (e.g., the robot is carrying the coffee too far above the table). In both of these cases the human is incentivized to physically interact with the robot and correct its behavior: but how should the robot respond? Here we formalize pHRI as a dynamical system where the robot does not know the correct objective function that the human wants it to optimize and the human's interactions are informative about this objective. Importantly, this formalism defines what it means for a robot to respond in the right or optimal way to physical human interactions. Furthermore, certain strategies for responding to pHRI can be justified as approximate solutions to this formalism. 


\medskip

\noindent \textbf{Notation}. Let $x$ be the robot's state, $u_r$ be the robot's action, and $u_h$ be the human's action. Returning to our motivating example, $x \in \mathbb{R}^n$ encodes the manipulator's joint positions and velocities, $u_r \in \mathbb{R}^m$ are the robot's commanded joint torques, and $u_h \in \mathbb{R}^m$ are the joint torques resulting from the wrench applied by the human. The robot transitions to the next state based on its deterministic dynamics ${\dot{x} = f(x,u_r + u_h)}$. Notice that both the robot's and human's action influence the robot's motion. In what follows we will work in discrete time, where a superscript $t$ denotes the current timestep. For instance, $x^t$ is the state at time $t$. \medskip

\noindent \textbf{Objective}. We model the human as having a particular reward function in mind that represents how they would like the current task to be performed. We write this reward function as a linear combination of task-related features \citep{abbeel2004apprenticeship,ziebart2008maximum}:
\begin{equation} \label{eq:F1}
    r(x,u_r,u_h ; \theta) = \theta \cdot \phi(x,u_r,u_h) - \lambda \|u_h \|^2.
\end{equation}
In the above, $\phi \in [0,1]^N$ is a normalized vector of $N$ features, $\lambda$ is a positive constant, and $\theta \in \mathbb{R}^N$ is a parameter vector that determines the relative weight of each feature. Here $\theta$ encapsulates the true \emph{objective}: if an agent knows exactly how to weight all the aspects of the task, then it can compute how to perform the task optimally. The first term in Equation (\ref{eq:F1}) is the task-related reward, while the second term penalizes human effort. Intuitively, the human wants the robot to complete the task according to their objective $\theta$---e.g., prioritizing keeping the coffee upright, or moving closer to the table---without any human intervention\footnote{We recognize that $\|u_h\|^2$ could also be thought of as a feature in $\phi$ with weight $\lambda$; however, we have explicitly listed this term to emphasize that the robot should not rely on human guidance.}.


With this formalism the robot should take actions $u_r$ to maximize the reward in Equation (\ref{eq:F1}) across every timestep. This is challenging, however, because the robot \emph{does not know} the true objective parameters $\theta$: only the human knows $\theta$. Different end-users have different objectives, which can change from task-to-task and even day-to-day. We thus think of $\theta$ as a hidden part of the state known only by the human. If the robot did know $\theta$, then pHRI would reduce to an instance of a Markov decision process (MDP), where the states are $x$, the actions are $u_r$, the reward is (\ref{eq:F1}), and the robot understands what it means to complete its task optimally. But since the actual robot is uncertain about $\theta$, we must reason over this uncertainty during pHRI.

\medskip


\noindent \textbf{POMDP}. We formalize pHRI as an instance of a partially observable Markov decision process (POMDP) where the true objective $\theta$ is a hidden part of the state, and the robot receives observations about $\theta$ through the human actions $u_h$. Formally, a POMDP is a tuple $\langle S, U, Z, T, O, r, \gamma \rangle$ where: 
\begin{itemize}
    \item $S$ is the set of states, where $s = (x,\theta)$, so that the system state contains the robot state $x$ and parameter $\theta$
    \item $U$ is the set of the robot actions $u_r$
    \item $Z$ is the set of observations (i.e. human actions $u_h$)
    \item $T(s^t, u^t_r + u^t_h, s^{t+1})$ is the transition distribution determined by the robot's dynamics ($\theta$ is constant)
    \item $O(s^{t+1}, u^t_r, z^{t+1})$ is the observation distribution
    \item $r(s^t, u^t_r, u^t_h)$ is the reward function from (\ref{eq:F1})
    \item $\gamma$ is the discount factor
\end{itemize}
In the above POMDP the robot cannot directly observe the system state $s$, and instead maintains a belief over $s$, where $b(s)$ is the probability of the system being in state $s$. Within our pHRI setting we assume that the robot knows its state $x$ (e.g., position and velocity), so that the belief over $s$ reduces to $b(\theta)$, the robot's belief over $\theta$. The robot does not know the human's true objective parameter $\theta$, but updates its belief over $\theta$ by observing the human's physical interactions $u_h$.

Solving this POMDP yields the robot's optimal response to pHRI during the task\footnote{The most general formulation for pHRI is that of a cooperative inverse reinforcement learning (CIRL) game \citep{hadfield2016cooperative}, which, when solved, yields the optimal human and robot policies.}. We point out that this POMDP is atypical, however, because the observations $u_h$ additionally affect the robot's reward $r$, similar to \cite{javdani2018shared}, and alter the robot's state $x$ via the transition distribution $T$. Because the human's actions can change both the state and reward, solving this POMDP suggests that the robot should anticipate future human actions, and choose control inputs $u_r$ that account for the predicted human inputs $u_h$, similar to \cite{hoffman2007cost}. \medskip

\noindent \textbf{Observation Model}. Assuming that human interactions are \emph{meaningful}, the robot should leverage the human's actions $u_h$ to update its belief over $\theta$. In order to associate the human interactions $u_h$ with the objective parameter $\theta$, the robot uses an observation model: $P(u_h \mid x,u_r; \theta)$. If we were to treat the human's actions as random disturbances, then we would select a uniform probability distribution for $P(u_h \mid x,u_r; \theta)$. By contrast, here we model the human as intentionally interacting to correct the robot's behavior; more specifically, let us model the human as correcting the robot to approximately maximize their reward. We assume the human selects an action $u_h$ that, when combined with the robot’s action $u_r$, leads to a high Q-value (state-action value) \emph{assuming the robot will behave optimally after the current timestep}, i.e., assuming that the robot learns the true $\theta$:
\begin{equation} \label{eq:F2}
    P(u^t_h \mid x^t,u^t_r; \theta) =  \frac{e^{Q(x^t,u^t_r+u^t_h; \theta)}}{\int e^{Q(x^t,u^t_r+\tilde{u}_h; \theta)} d\tilde{u}_h}
\end{equation}
Our choice of Equation (\ref{eq:F2}) stems from maximum entropy assumptions \citep{ziebart2008maximum}, as well as the Bolzmann distributions used in cognitive science models of human behavior \citep{baker2007goal}. 

\section{Approximate Solutions for Online Learning} \label{sec:approximation}


Although we have demonstrated that pHRI is an instance of a POMDP, solving POMDPs exactly is at best computationally expensive and at worst intractable \citep{kaelbling1998planning}. POMDP solvers have made significant progress \citep{silver2010pomcp, somani2013despot}; however, it still remains difficult to compute online solutions for continuous state, action, and observation spaces. For instance, when evaluated on a toy problem ($S = \mathbb{R}^4$, $O = \mathbb{R}^8$), recent developments do not obtain exact solutions within one second \citep{sunberg2017pomcpow}. The lack of efficient POMDP solvers for large, continuous state, action, and observation spaces is particularly challenging here since (a) the dimension of our state space $S$ is twice the number of robot DoF, $2n$, plus the number of task-related features, $N$, and (b) we are interested in real-time solutions that enable the robot to learn and act while the human is interacting (i.e. we need millisecond-to-second solutions). Accordingly, in this section we introduce three approximations to our pHRI formalism that enable online solutions. First, we separate finding the optimal robot policy from estimating the human's objective. Next, we simplify the observation model and use a maximum \textit{a posteriori} (MAP) estimate of $\theta$ as opposed to the full belief over $\theta$. Finally, when finding the optimal robot policy and estimating $\theta$, we move from \emph{policies to trajectories}. These approximations show how our solution is derived from the complete POMDP formalism outlined in the last section, but now enable the robot to learn and react in real-time with continuous state, action, and belief spaces.

\medskip

\noindent \textbf{QMDP}. We first assume that $\theta$ will become fully observable to the robot at the next timestep. Given this assumption, our POMDP reduces to a QMDP \citep{littman1995learning}; QMDPs have been used by \cite{javdani2018shared} to approximate a POMDP with uncertainty over the human's goal. The QMDP separates into two distinct subproblems: (a) \emph{finding the robot's optimal policy} given the current belief $b(\theta)$ over the human's objective:
\begin{equation} \label{eq:A1}
    Q(x,u_r,b) = \int b(\theta) Q(x,u_r,\theta) d\theta
\end{equation}
where $u^*_r = \text{arg}\max_{u_r} Q(x,u_r,b)$ evaluated at every state yields the optimal policy, and (b) \emph{updating the belief $b(\theta)$ over the human's objective $\theta$} given a new observation:
\begin{equation} \label{eq:A1p2}
b^{t+1}(\theta) = \frac{P(u^t_h \mid x^t, u^t_r; \theta) b^t(\theta)}{\int P(u^t_h \mid x^t, u^t_r; \tilde{\theta})b^t(\tilde{\theta})d\tilde{\theta}}   
\end{equation}
where $P(u^t_h \mid x^t, u^t_r; \theta)$ is the observation model in Equation (\ref{eq:F2}), and $b^t(\theta) = P(\theta \mid x^{0:t}, u^{0:t}_r, u^{0:t}_h)$ for $t \in \{0,1,\hdots\}$.

Intuitively, under this QMDP the robot is always exploiting the information it currently has, and never actively tries to explore for new information. A robot using the policy from Equation~(\ref{eq:A1}) does not anticipate any human actions $u_h$, and so the robot solves for its optimal policy as if it were completing the task in isolation. Recall that we previously pointed out that physical human interactions can influence the robot's state. In practice, however, we do not necessarily want to account for these actions when planning---the robot should not rely on the human to move the robot. Due to our QMPD approximation the robot never relies on the human for guidance: but when the human does interact, the robot leverages $u_h$ to learn about $\theta$ in Equation (\ref{eq:A1p2}). In summary, the robot only considers $u_h$ for its information value.


\medskip

\noindent \textbf{MAP of $\theta$}. Ideally, the robot would maintain a full belief $b(\theta)$ over $\theta$. Since the human's objective $\theta \in \mathbb{R}^N$ is continuous, potentially high-dimensional, and our observation model is non-Gaussian, we approximate $b$ with the maximum \textit{a posteriori} estimate. We will let $\hat{\theta}$ be the robot's MAP estimate of $\theta$.

\medskip

\noindent \textbf{Planning and Control}. Indeed, even if we had $b(\theta)$, solving (\ref{eq:A1}) in continuous state, action, and belief spaces is still intractable for real-time implementations. Let us focus on the challenge of finding the robot's optimal policy given the current MAP estimate $\hat{\theta}$. We move from computing policies to planning trajectories, so that---rather than evaluating (\ref{eq:A1}) at every timestep---we \emph{plan} an optimal trajectory from start to goal, and then track that trajectory using a safe \emph{controller}.

At every timestep $t$, we first \emph{replan} a trajectory $\xi = x^{0:T} \in \Xi$ which optimizes the task-related reward from Equation (\ref{eq:F1}) over the $T$-step planning horizon. If our features $\phi$ only depend on the state $x$, then the cumulative task-related reward becomes:
\begin{equation} \label{eq:A2}
    R(\xi ; \theta) = \theta \cdot \Phi(\xi) = \sum_{x^t \in \xi} \theta \cdot \phi(x^t)
\end{equation}
Here $\Phi(\xi)$ is the total feature count along trajectory $\xi$. Using the cumulative reward function in Equation (\ref{eq:A2}), the robot finds the optimal trajectory $\xi_r^t$ from its current estimate $\hat{\theta}^t$:
\begin{equation} \label{eq:A3}
    \xi_r^t = \text{arg}\max_{\xi \in \Xi} \hat{\theta}^t \cdot \Phi(\xi)
\end{equation}
We can solve Equation (\ref{eq:A3}) for the optimal trajectory using trajectory optimization tools \citep{schulman2014motion, karaman2011sampling}. Whenever $\hat{\theta}$ is updated from pHRI during task execution, the robot's trajectory will be replanned using that new estimate to match the the learned objective.

To track the robot's planned trajectory we leverage \emph{impedance control}. Impedance control---as originally proposed by \cite{hogan1985impedance}---is the most popular controller for pHRI \citep{haddadin2016physical}, and ensures that the robot responds compliantly to human corrections \citep{de2006collision}. Let $x^t = (q^t, \dot{q}^t)$, where $q^t$ is the robot's current configuration, and $q^t_r \in \xi^t_r$ is the desired configuration at timestep t. After feedback linearization \citep{spong2006robot}, the equation of motion of a robot arm under impedance control becomes:
\begin{equation} \label{eq:A4}
    M_r(\ddot{q}^t - \ddot{q}^t_r) + B_r(\dot{q}^t - \dot{q}_r^t) + K_r(q^t - q_r^t) = u_h^t
\end{equation}
Here $M_r$, $B_r$, and $K_r$ are the desired inertia, damping, and stiffness rendered by the robot. These parameters determine what impedance the human perceives: for instance, lower $K_r$ makes the robot appear more compliant. In our experiments, we implement a simplified impedance controller without feedback linearization:
\begin{equation} \label{eq:A5}
    u_r^t = B_r(\dot{q}^t_r - \dot{q}^t) + K_r(q^t_r - q^t)
\end{equation}
This control input drives the robot towards its desired state $x^t \in \xi_r^t$, and evaluating Equation (\ref{eq:A5}) over all states yields the robot's policy. To summarize, we first solve the trajectory optimization problem from Equation (\ref{eq:A3}) to get the current robot trajectory $\xi_r^t$, and then compliantly track that trajectory using Equation (\ref{eq:A5}). Notice that if the robot never updates $\hat{\theta}$ then $\xi_r^t = \xi_r^{t-1}$, and this approach reduces to using impedance control to track an unchanging robot trajectory. \medskip

\noindent \textbf{Intended Trajectories}. Next we address the second QMDP subproblem: updating the MAP estimate $\hat{\theta}$ after each new observation. First we must find an observation model which we can compute in real-time. Similar to solving for our optimal policy with Equation (\ref{eq:A1}), evaluating our observation model from Equation (\ref{eq:F2}) for a given $\theta$ is challenging because it requires that we determine the $Q$-value associated with that $\theta$. Previously we avoided this issue by moving from policies to trajectories. We will utilize the same simplification here to find a feasible observation model based on the human's intended trajectory.

Instead of attempting to directly relate $u_h$ to $\theta$, as in our original observation model, we propose an intermediate step: interpret each human action $u_h$ via an \emph{intended trajectory}, $\xi_h$, which the human would prefer for the robot to execute. We leverage trajectory deformations \citep{losey2018trajectory} to get the intended trajectory $\xi_h$ from the robots planned trajectory $\xi_r$ and the humans physical interaction $u_h$. Following \cite{losey2018trajectory}, we propagate the human's interaction force along the robot's trajectory:
\begin{equation} \label{eq:A6}
    \xi_h = \xi_r + \mu A^{-1} U_h
\end{equation}
where $\mu > 0$ scales the magnitude of the deformation. The symmetric positive definite matrix $A$ defines a norm on the Hilbert space of trajectories and dictates the shape of the deformation \cite{dragan2015movement}. The input vector is $U_h = u_h$ at the current time, and $U_h=0$ at all other times. During experiments we use the velocity norm for $A$ \citep{dragan2015movement}, but other options are possible.

\textcolor{black}{Our deformed trajectory minimizes the distance from the previous trajectory while keeping the end-points the same and moving the corrected point to its new configuration \citep{dragan2015movement}. Whereas using the Euclidean norm to measure distance would return the same trajectory as before with the current waypoint teleported to where the user corrected it, using a band-diagonal norm $A$ (e.g., the velocity norm) serves to couple each waypoint along the trajectory to the one before it and the one after it. This formalizes the effect proposed by elastic strips by \cite{brock2002elastic} and elastic bands by \cite{quinlan1993elastic}.}

Now rather than evaluating the $Q$-value of $u_h + u_r$ given $\theta$, like we did in Equation (\ref{eq:F2}), we can compare the human's intended trajectory $\xi_h$ to the robot's original trajectory $\xi_r$ and relate these differences to $\theta$. We assume that the human provides a intended trajectory $\xi_h$ that approximately maximizes their cumulative task-related reward from Equation (\ref{eq:A2}) while remaining close to $\xi_r$:
\begin{align} \label{eq:A7}
    P(\xi_h \mid \xi_r; \theta) 
    &\approx \frac{e^{R(\xi_h ; \theta) - \lambda \|\xi_h - \xi_r\|^2}}{\int e^{R(\tilde{\xi}_h ; \theta) - \lambda \|\tilde{\xi}_h - \xi_r\|^2} d\tilde{\xi}_h}
\end{align}
Moving forward we treat $P(\xi_h \mid \xi_r; \theta)$ as our observation model. Note that this observation model is analogous to Equation (\ref{eq:F2}) but in trajectory space. In other words, Equation (\ref{eq:A7}) yields a distribution over intended trajectories given $\theta$ and the current robot trajectory. Here the correspondence between the human's effort $\|u_h\|^2$ and the change in trajectories $\|\xi_h - \xi_r\|^2$ stems from the deformation in Equation (\ref{eq:A6}). In conclusion, we can leverage our simplified observation model (\ref{eq:A7}) to tractably reason about the meaning behind the human's physical interaction. \section{All-at-Once Online Learning} \label{sec:all}

So far we have determined how to choose the robot's actions given $\hat{\theta}$, the current MAP estimate of the human's objective. We have also derived a tractable observation model. Next, we apply this observation model to update $\hat{\theta}$ based on human interactions. By using online gradient descent we arrive at an update rule for $\hat{\theta}$ which adjusts the weights of all the features based on a single human correction. We refer to this method as \emph{all-at-once} learning. We also relate all-at-once learning to prior works on online Maximium Margin Planning (MMP) and Coactive Learning. \medskip

\noindent \textbf{Gradient Descent}. If we assume that the observations are conditionally independent\footnote{Recent work by \cite{li2021learning} extends our approach to cases where the interactions are not conditionally independent, i.e., multiple corrections are interconnected.}, then the maximum \textit{a posteriori} (MAP) estimate at timestep $t+1$ is:
\begin{align} \label{eq:L1}
    \hat{\theta}^{t+1} &= \text{arg}\max_{\theta} P(\xi_h^0,\ldots,\xi_h^t \mid \xi_r^0,\ldots,\xi_r^t, \theta)P(\theta) \nonumber \\ &= \text{arg}\max_{\theta} \sum_{\tau = 0}^t \ln{P(\xi_h^{\tau} \mid \xi_r^{\tau}, \theta)} + \ln{P(\theta)}
\end{align}
where $P(\xi^{\tau}_h \mid \xi^{\tau}_r; \theta)$ is our observation model from Equation (\ref{eq:A7}). To use this model we need to compute the normalizer, which requires integrating over the space of all possible human-preferred trajectories. We instead leverage Laplace's method to approximate the normalizer. Taking a second-order Taylor series expansion of $R(\xi_h , \theta) - \lambda \|\xi_h - \xi_r\|^2$ about $\xi_r$, the robot's estimate of the optimal trajectory, we obtain a Gaussian integral that we can evaluate:
\begin{equation} \label{eq:L2}
    P(\xi_h \mid \xi_r, \theta) \approx e^{R(\xi_h , \theta) - R(\xi_r , \theta) - \lambda \|\xi_h - \xi_r\|^2}
\end{equation}
Since we have assumed that the human's intended trajectory $\xi_h$ is an \emph{improvement} over the robot's trajectory $\xi_r$, then  it must be the case that $R(\xi_h , \theta) > R(\xi_r , \theta)$. Let $\hat{\theta}^0$ be the robot's initial estimate of $\theta$, such that the robot has a prior:
\begin{equation} \label{eq:L3}
    P(\theta) = \frac{1}{(2\pi\alpha)^{1/2}}e^{-\frac{1}{2\alpha}\|\theta - \hat{\theta}^0\|^2}
\end{equation} 
where $\alpha$ is a positive constant. 

Substituting our normalized observation model from Equation (\ref{eq:L2}) and the prior from Equation (\ref{eq:L3}) back into Equation (\ref{eq:L1}), the MAP estimate $\hat{\theta}^{t+1}$ is the solution to:
\begin{equation} \label{eq:L4}
    \text{arg}\max_{\theta} \sum_{\tau = 0}^t \Big(R(\xi_h^{\tau} , \theta) - R(\xi_r^{\tau} , \theta)\Big) - \frac{1}{2\alpha}\|\theta - \hat{\theta}^0\|^2
\end{equation}
In Equation (\ref{eq:L4}) the $\lambda \|\xi_h - \xi_r\|^2$ terms have dropped out because this penalty for human effort does not explicitly depend on $\theta$. Intuitively, our estimation problem (\ref{eq:L4}) states that we are searching for the objective $\theta$ that \emph{maximally separates} the reward associated with $\xi_h$ and $\xi_r$, while also regulating the size of the change in $\theta$.

We solve Equation (\ref{eq:L4}) by taking the gradient with respect to $\theta$ and then setting the result equal to zero. Substituting in our cumulative reward function from Equation (\ref{eq:A2}), we obtain the \emph{all-at-once} update rule:
\begin{align} \label{eq:L5}
    \hat{\theta}^{t+1} &= \hat{\theta}^0 + \alpha \sum_{\tau = 0}^t \big(\Phi(\xi_h^{\tau}) - \Phi(\xi_r^{\tau})\big) \nonumber \\ &= \hat{\theta}^t + \alpha \big(\Phi(\xi_h^t) - \Phi(\xi_r^t)\big)
\end{align}
Given the current MAP estimate $\hat{\theta}^t$, the robot's trajectory $\xi_r^t$, and the human's intended trajectory $\xi_h^t$, we determine an approximate MAP estimate at timestep $t+1$ by comparing the feature counts. Note that the update rule in (\ref{eq:L5}) is actually the online gradient descent algorithm \citep{bottou1998online} applied to our normalized observation model (\ref{eq:L2}).

\medskip

\begin{figure}[!t]
\centering
\includegraphics[width=.5\textwidth]{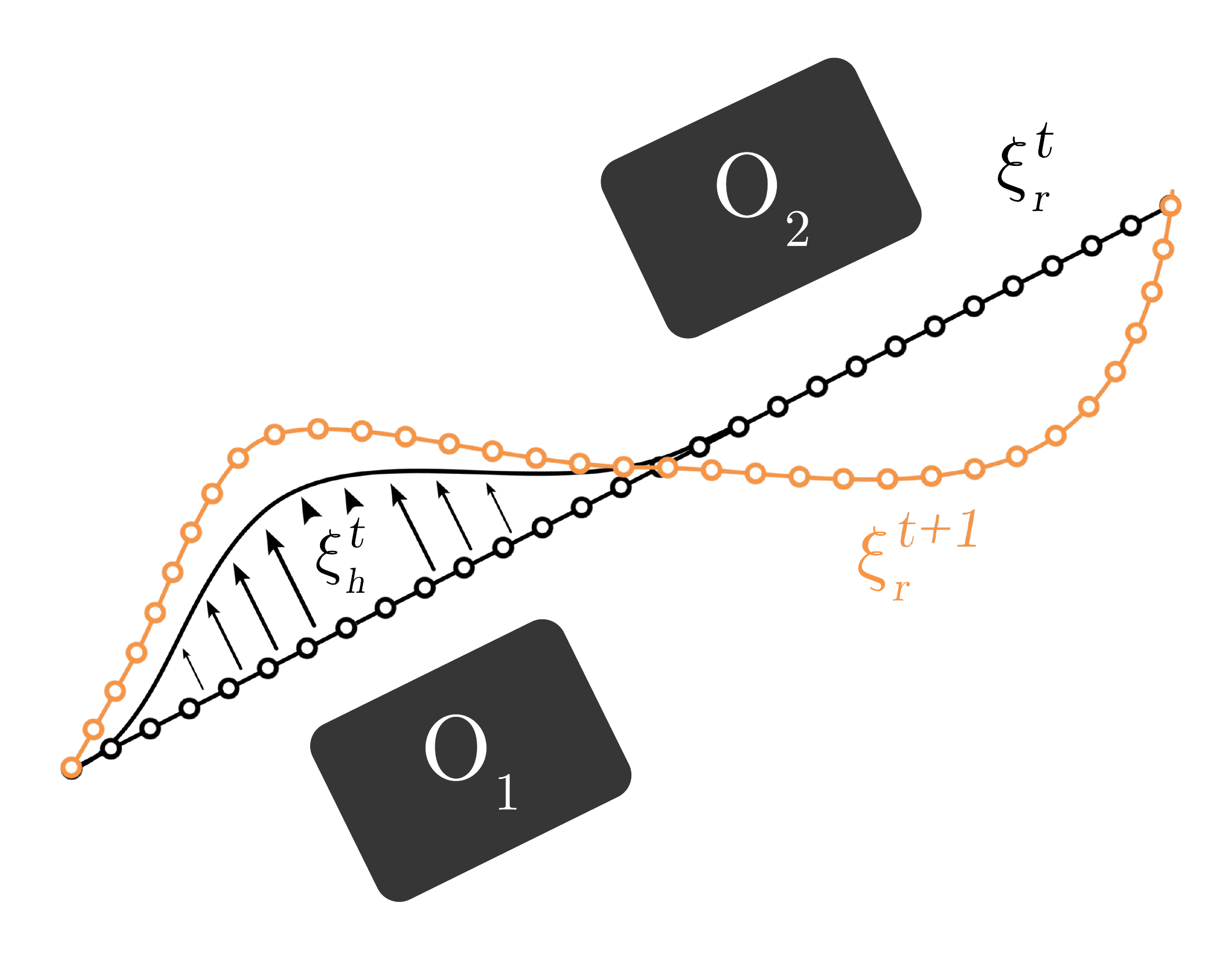}
\caption{Visualization of one iteration of our proposed algorithm for online learning from pHRI. Here a point robot is moving in a $2$D environment with two obstacles, $O_1$ and $O_2$. The robot initially plans to follow a straight line trajectory from start to goal ($\xi^{t}_r$, black dotted line). But the human wants the robot to move farther away from the obstacles: the human pushes the robot, and the robot uses the human's applied force to deform its initial trajectory into a human preferred trajectory ($\xi^{t}_h$, solid black line). Given that $\xi^{t}_h$ is better aligned with the human's objective than $\xi^{t}_r$, we compute an online update of $\theta$ and replan a new trajectory $\xi^{t+1}_r$ (orange dotted line). Notice that the new trajectory moves the robot farther from the nearby obstacle $O_1$ and the future obstacle $O_2$.}
\label{fig:algorithm}
\vspace{-1em}
\end{figure}

\noindent \textbf{Interpretation}. The all-at-once update rule (\ref{eq:L5}) has a simple interpretation: if any feature has a higher value along the human's intended trajectory than the robot's trajectory, the robot should increase the weight of that feature. Returning to our example, if the human's preferred trajectory $\xi_h$ moves the coffee closer to the table than the robot's original trajectory $\xi_r$, the weights in $\hat{\theta}$ for distance-to-table will increase. This enables the robot to learn in real-time from corrections.

Interestingly, our all-at-once update rule is a special case of the update rules from two related IRL works. Equation (\ref{eq:L5}) is the same as the Preference Perceptron for coactive learning---introduced in \cite{shivaswamy2015} and applied for manipulation tasks by \cite{jain2015learning}---if $\xi_h$ was the robot's original trajectory $\xi_r$ with a single corrected waypoint. Similarly, Equation (\ref{eq:L5}) is analogous to online Maximum Margin Planning without the loss function if the correction $\xi_h$ was treated as a new demonstration \citep{ratliff2006maximum}. These findings also align with work from  \cite{choi2011map}, who show that other IRL methods can be interpreted as a MAP estimate. What is unique in our work is that we demonstrate how the online gradient-descent update rule in Equation (\ref{eq:L5}) results from a POMDP with hidden state $\theta$ where physical human interactions are interpreted as intended trajectories.

 \section{One-at-a-Time Online Learning} \label{sec:one}

We derived an update rule to learn the human's objective from their physical interactions with the robot. This all-at-once approach changes the weight of \emph{all} the features that the human adjusts during their correction. In practice, however, the human's interactions (and their intended trajectory) may result in \emph{unintended corrections} which mistakenly alter features the human meant to leave untouched. For example, when the human's action intentionally causes $\xi_h$ to move closer to the table, the same correction may accidentally also change the orientation of the coffee. In order to address unintended corrections, we here assume that the human's intended trajectory $\xi_h$ should change only a \emph{single} feature. We explain how to determine which feature the human is trying to change, and then modify the update rule from Equation (\ref{eq:L5}) to obtain \emph{one-at-a-time} learning. 

\medskip

\noindent \textbf{Intended Feature Difference}. Let us define the change in features at time $t$ as $\Delta \Phi^t = \Phi(\xi_h^t) - \Phi(\xi_r^t) \in \mathbb{R}^N$, where $\xi^t_h$ is the human's intended trajectory, $\xi^t_r$ is the robot's trajectory, and $N$ is the number of features. Given our assumption that the human intends to change just one feature at a single timepoint, $\Delta \Phi^t$ should have only a single non-zero entry; however, because human corrections are imperfect \citep{akgun2012keyframe,jonnavittula2020know} this not always the case. We introduce the \emph{intended feature difference}, $\Delta \Phi^t_h$, where only the feature the human wants to update is non-zero. At each timestep the robot must infer $\Delta \Phi^t_h$ from $\Delta \Phi^t$. Note that this one-at-a-time approach does not mean that only a single feature changes during the entire task: the user can adjust a \emph{different} feature at each timestep.

Without loss of generality, assume the human is trying to change the $i$-th entry of the robot's MAP estimate $\hat{\theta}$ during the current timestep $t$. The ideal human correction of $\xi_r^t$ should accordingly change the feature count in the direction:
\begin{equation} \label{eq:O1}
    J_i = \frac{\partial \Phi(\xi_r^t)}{\partial \hat{\theta}^t_i}
\end{equation}
Recall that $\xi_r^t$ is optimal with respect to the current estimate $\hat{\theta}^t$, and so changing $\hat{\theta}^t$ will alter $\Phi(\xi_r^t)$. Put another way, if the human is an optimal corrector, and their interaction was meant to alter just the weight on the $i$-th feature, then we would expect them to correct the current robot trajectory $\xi_r^t$ such that they produce a feature difference $\Delta \Phi^t$ exactly in the direction of the vector $J_i$ from Equation (\ref{eq:O1}).

Because the human is imperfect, they will not exactly match Equation (\ref{eq:O1}). Instead, we model the human as making corrections $\Delta \Phi^t$ in the direction of $J_i$. This yields an \emph{observation model} from which the robot can find the likelihood of observing a specific feature difference $\Delta \Phi^t$ given that the human is attempting to update the $i$-th feature:
\begin{equation} \label{eq:O2}
    P(\Delta \Phi 
    \mid i) \propto e^{|J_i \cdot \Delta \Phi|}
\end{equation}
Recalling that the robot observes the feature difference $\Delta \Phi^t = \Phi(\xi_h^t) - \Phi(\xi_r^t)$, then we estimate which feature the human most likely wants to change using:
\begin{align} \label{eq:O3}
    i^* &= \text{arg}\max_{i} P(\Phi(\xi_h^t) - \Phi(\xi_r^t) \mid i) \nonumber \\ &= \text{arg}\max_{i} \big|J_i \cdot \big(\Phi(\xi_h^t) - \Phi(\xi_r^t)\big)\big|
\end{align}
Once the robot solves for the most likely feature the human wants to change, $i^*$, it can now find the human's \emph{intended feature difference} $\Delta \Phi^t_h$. Recall that, if the human wanted to only update feature $i^*$, their intended feature difference would ideally be in the direction $J_{i^*}$. Thus, we choose $\Delta \Phi^t_h \propto J_{i^*}$ as our intended feature difference. 

\medskip

\noindent \textbf{Update Rule}. We make two simplifications to derive a one-at-a-time update rule. Both simplifications stem from the difficulty of evaluating the partial derivative from Equation (\ref{eq:O1}) in real-time. Indeed, rather than computing this partial derivative, we approximate $J_i$ as proportional to the vector $(0,\ldots,1,\ldots,0)$, where the $i$-th entry is non-zero. Intuitively, we are here assuming that when the $i$-th weight in $\hat{\theta}$ changes, it predominately induces a change in the $i$-th feature along the resulting optimal trajectory. 

Given this assumption, computing the intended feature difference $\Delta \Phi^t_h \propto J_{i^*}$ reduces to projecting the observed feature difference $\Delta \Phi^t$ induced by the human's action $u_h$ onto the $i^*$-th axis:
\begin{equation} \label{eq:O4}
    \Delta \Phi^t_h = (0,\ldots,\Delta\Phi^t_{i^*},\ldots,0)
\end{equation}
This fulfills our original requirement for the intended feature difference $\Delta \Phi^t_h$ to only have one non-zero entry. Moreover, once we substitute our simplification of $J_i$ back into our feature estimation problem (\ref{eq:O3}), we get a simple yet intuitive heuristic for finding $i^*$: only the feature which the user has changed the \emph{most} during their correction should be updated. Our \emph{one-at-a-time} update rule is therefore similar to the gradient update from Equation (\ref{eq:L5}), but with a single feature weight update using Equation (\ref{eq:O4}):
\begin{equation} \label{eq:O5}
    \hat{\theta}^{t+1} = \hat{\theta}^t + \alpha \Delta \Phi_h^t
\end{equation}
Instead of updating the estimated weights associated with all the features like in Equation (\ref{eq:L5}), we now only update the MAP estimate for the feature which has the largest change in feature count. Overall, isolating a single feature at every timestep is meant to mitigate the effects of unintended learning from noisy physical interactions\footnote{We note that all the features are normalized to have the same sensitivity.}. \section{Optimally Responding to pHRI} 

Before introducing all-at-once and one-at-a-time learning, we showed how approximate solutions to pHRI involve (a) safely tracking the optimal trajectory and (b) updating the MAP estimate based on human interactions. Now that we have derived update rules for $\hat{\theta}$, we will circle back and present our algorithm for learning from pHRI. We also include practical considerations for implementation. 

\medskip 

\noindent \textbf{Algorithm}. We have formalized pHRI as an instance of a POMDP and then approximated that POMDP as a QMDP. To solve this QMDP we must both find the robot's optimal policy and update the MAP estimate of $\theta$ at every timestep $t$. First, we approximate the robot's optimal policy by solving a trajectory optimization problem in Equation (\ref{eq:A3}) for $\xi_r^t$ and then tracking $\xi_r^t$ with an impedance controller (\ref{eq:A5}). Second, we update the MAP estimate $\hat{\theta}^t$ by interpreting each human correction as an intended trajectory---which we obtain by deforming the robot's original trajectory using Equation (\ref{eq:A6})---and next we perform either all-at-once (\ref{eq:L5}) or one-at-a-time (\ref{eq:O5}) online updates to obtain $\hat{\theta}^{t+1}$. At the next timestep $t+1$ the robot replans its optimal trajectory under $\hat{\theta}^{t+1}$ and the process repeats. An overview is provided in Algorithm~\ref{alg:a1}.

\begin{algorithm} [t!]
	\caption{Online Learning from pHRI}
	\label{alg:a1}
	\begin{algorithmic}
	\setstretch{1.2}

\State {Given: initial weights $\hat{\theta}^0$ and features $\phi \in [0,1]^N$}
\State {Initialize: $\xi_r^0 \gets \text{arg}\max_{\xi} \hat{\theta}^0 \cdot \Phi(\xi)$}

	\setstretch{1.5}
\For {$t = 0$ to $T$}
	\State $u_r^t = B_r(\dot{q}^t_r - \dot{q}^t) + K_r(q^t_r - q^t)$ \Comment{(\ref{eq:A5})}
	\State $\xi_h^t \gets \xi_r^t + \mu A^{-1} U_h^t$ \Comment{(\ref{eq:A6})}
	\State $\hat{\theta}^{t+1} \gets \hat{\theta}^t + \alpha \big(\Phi(\xi_h^t) - \Phi(\xi_r^t)\big)$ \Comment{(\ref{eq:L5}) or (\ref{eq:O5})}
	\State $\xi_r^{t+1} \gets \text{arg}\max_{\xi} \hat{\theta}^{t+1} \cdot \Phi(\xi)$ \Comment{(\ref{eq:A3})}
\EndFor

	\end{algorithmic}

\end{algorithm}

\medskip

\noindent \textbf{Implementation}. In practice, Algorithm~\ref{alg:a1} uses impedance control to track a trajectory that is replanned after pHRI. We note, however, that this approach ultimately derives from formulating pHRI as a POMDP. One possible variation on this algorithm is---instead of replanning $\xi_r^t$ from start to goal---replanning $\xi_r^t$ from the robot's current state $x^t$ to the goal. The advantage of this variation is that it saves us the time of recomputing the trajectory before our current state (which the robot does not need to know). However, in our implementation we always replan from start to goal. This is because constantly setting $x^t$ along the desired trajectory prevents the human from experiencing any impedance during interactions (i.e., the robot never resists the human's interactions). Without any haptic feedback from the robot, the end-user cannot easily infer the current robot's trajectory, and so the human does not know whether additional corrections are necessary \citep{jarrasse2012framework}. A second consideration deals with the robot's feature space. Throughout this work we assume that the robot knows the relevant features $\phi$, which are provided by the robot designer or user \citep{argall2009survey}. Alternatively, the robot could use techniques like feature selection \citep{guyon2003introduction} to filter a set of available features, or the features could be learned by the robot \citep{levine2016end}.
\section{Simulations}

To compare our real-time learning approach with optimal offline solutions and current online baselines, as well as to test both all-at-once and one-at-a-time learning, we conduct human-robot interaction simulations in a controlled environment. Here the robot is performing a pick-and-place task: the robot is carrying a cup of coffee for the simulated human. The simulated human physically interacts with the robot to correct its behavior. 

\medskip

\noindent \textbf{Setup}. We perform three separate simulated experiments. In each, the robot is moving within a planar world from a fixed start position to a fixed goal position. We here use a 2-DoF \emph{point robot} for simplicity, while noting that we will use a 7-DoF robotic manipulator during our user studies. The robot's state is $x \in \mathbb{R}^2$, the robot's action is $u_r\in \mathbb{R}^2$, and the human's action is $u_h \in \mathbb{R}^2$; both the state and action spaces are continuous. We assume that the robot knows the relevant features $\phi$, but the robot does not know the human's objective $\theta$. The robot initially believes that ``velocity'' (i.e., trajectory length) is the only important feature, and so the robot tries to move in a \emph{straight line} from start to goal.
\medskip

\begin{figure}[t]

	\begin{center}
		\includegraphics[width=0.8\columnwidth]{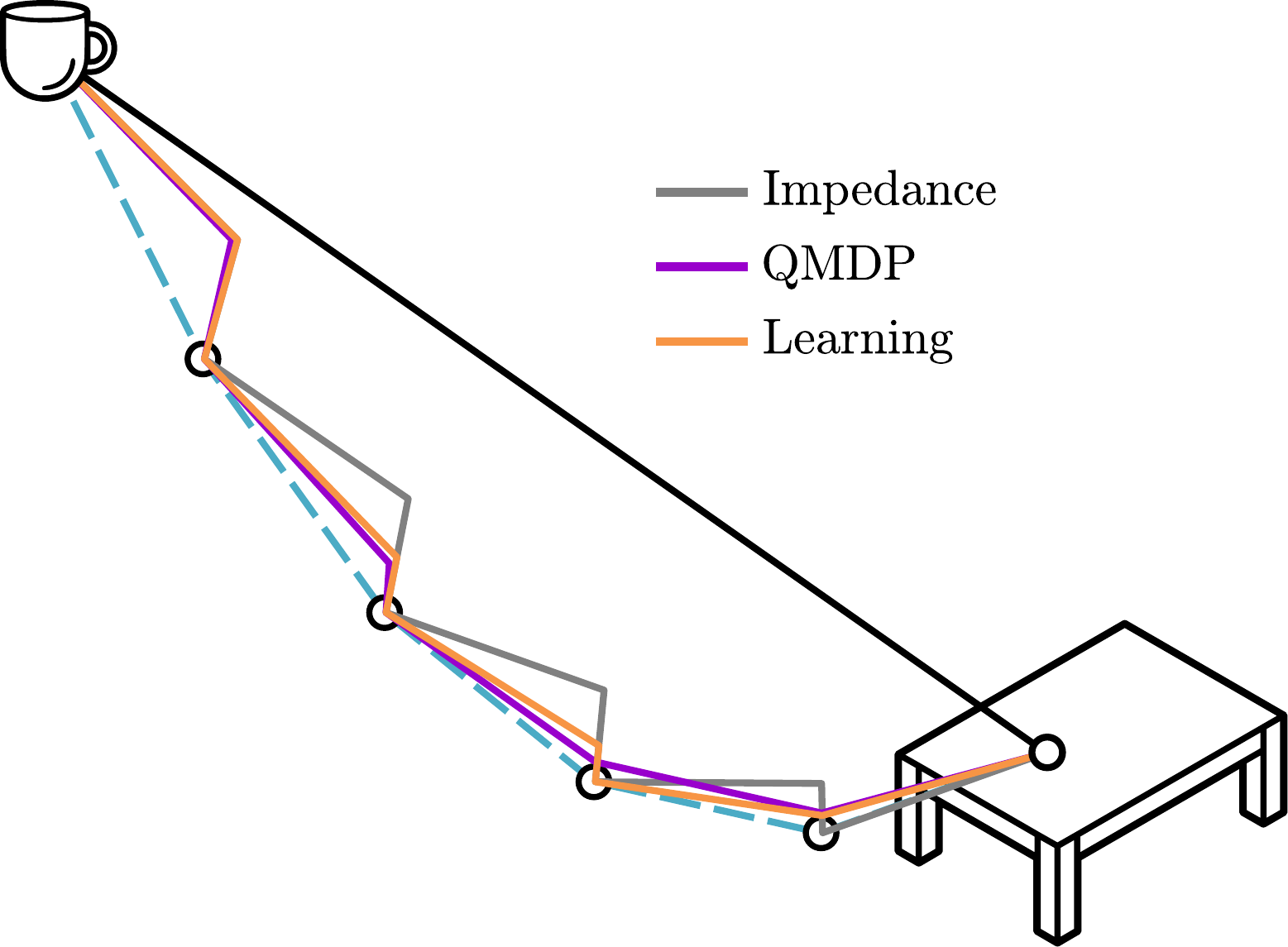}

		\caption{Comparison of the offline QMDP solution and our online Learning approximation for a pick-and-place task. The robot is attempting to carry the cup to the table. Originally the robot is confident it should move in a straight line (black), but the user actually wants the cup to be carried closer to the ground (blue, dashed). Here the human physically interacts to guide the robot back to their desired trajectory (circles) when the robot's error is too high.}

		\label{fig:sim1_task}
	\end{center}

	\vspace{-0.5em}

\end{figure}

\begin{figure}[t]

	\begin{center}
		\includegraphics[width=1\columnwidth]{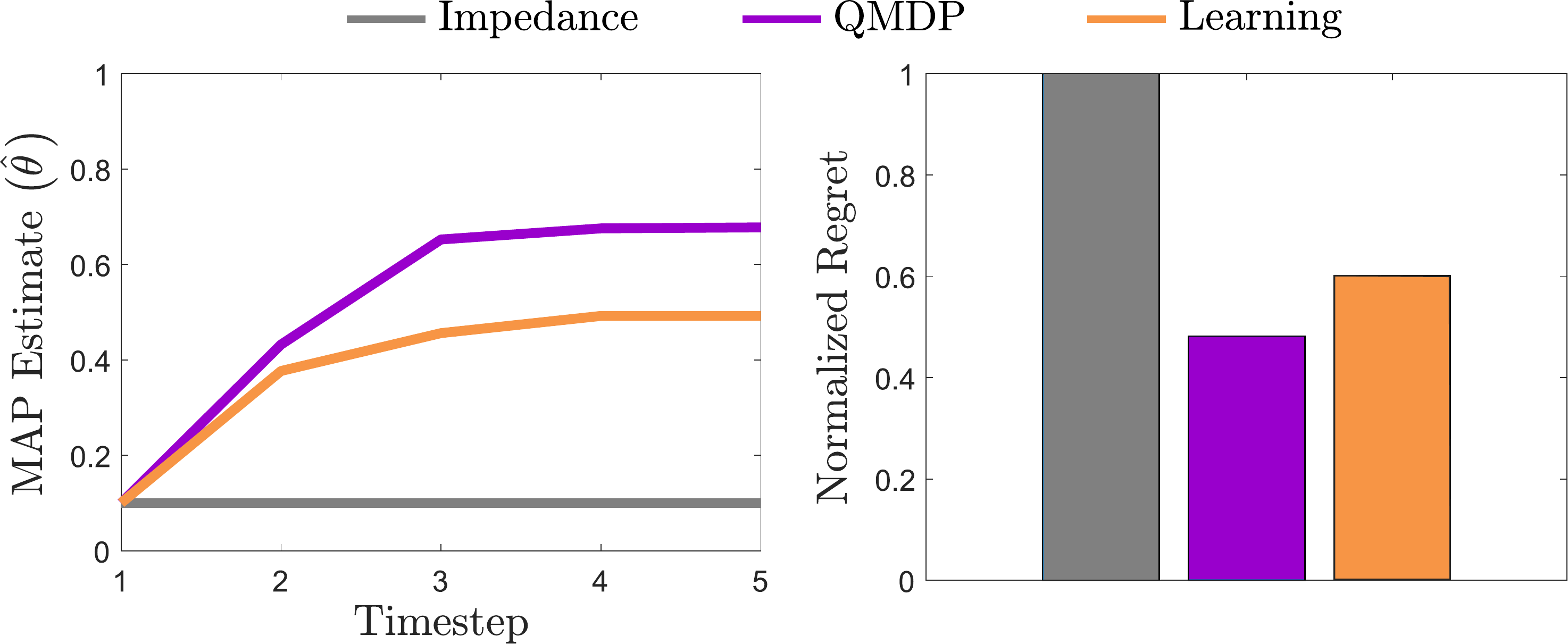}

		\caption{Robot learning and regret for the task from Fig.~\ref{fig:sim1_task}. The true human objective is $\theta=1$. The offline QMDP solution learns more about the human's objective than our online Learning approximation. However, both QMDP and Learning lead to significantly less regret than the Impedance baseline. The regret for QMDP is the lowest because here human corrects the robot at one less timestep.}

		\label{fig:sim1_metric}
	\end{center}

	\vspace{-1em}

\end{figure}


\noindent \textbf{Learning vs. QMDP vs. No Learning}. To learn in real-time, we introduced several approximations on top of separating estimation from control (QMDP). Here we want to assess how much these approximations reduce the robot's performance. We first compare our approximate real-time solution described in Algorithm~\ref{alg:a1} to the complete QMDP solution \citep{littman1995learning}. As a baseline, we have also included just using impedance control \citep{haddadin2016physical}, where no learning takes places from the humans interactions. Thus, the three tested approaches are \emph{Impedance}, \emph{QMDP}, and \emph{Learning}. The simulated task is depicted in Fig.~\ref{fig:sim1_task}. The two features are ``velocity'' and ``table,'' and the human wants the robot to carry their coffee closer to table level ($\theta=1$). During each timestep, if the robot's position error from the human's desired trajectory exceeds a predefined threshold, then the human physically corrects the robot by guiding it to their desired trajectory. Recall that our Learning method uses a MAP estimate of the human's objective, but the full QMDP solution maintains a belief $b$ over $\theta$. For QMDP simulations, we discretize the belief space---such that $\theta \in \{0,1\}$---and the robot starts with a prior $b^0(\theta=1) = 0.1$. Using a planar environment and a discretized belief space enables us to actually compare the full QMDP solution to our approximation, since the QMDP becomes prohibitively expensive in high dimensions with continuous state, action, and belief spaces.

We expect the full QMDP solution to outperform our Learning approximation. From Fig.~\ref{fig:sim1_metric}, we observe that the robot learns $\theta$ faster when using the QMDP, and that the robot completes the task with less regret. Both QMDP and Learning outperform Impedance, where the robot does not learn from pHRI. We note that here the simulated human behaves differently than our observation model (\ref{eq:F2}): rather than maximizing their $Q$-value, the human is guiding the robot along their desired trajectory. When the simulated human \emph{does} follow our observation model, we obtain very similar results: the normalized regret becomes $0.55$ for QMDP and $0.62$ for Learning. To ensure that the learning rate is consistent between the QMDP and Learning methods, we selected $\alpha$ such that $\hat{\theta}^1$ equalled $b^1(\theta = 1)$ when the simulated human followed our observation model (\ref{eq:F2}). From these simulations we conclude that the Learning approximation for online performance is worse than the full QMDP solution, but the difference between these methods is \emph{negligible} when compared to Impedance.

\medskip

\begin{figure}[t]

	\begin{center}
		\includegraphics[width=0.8\columnwidth]{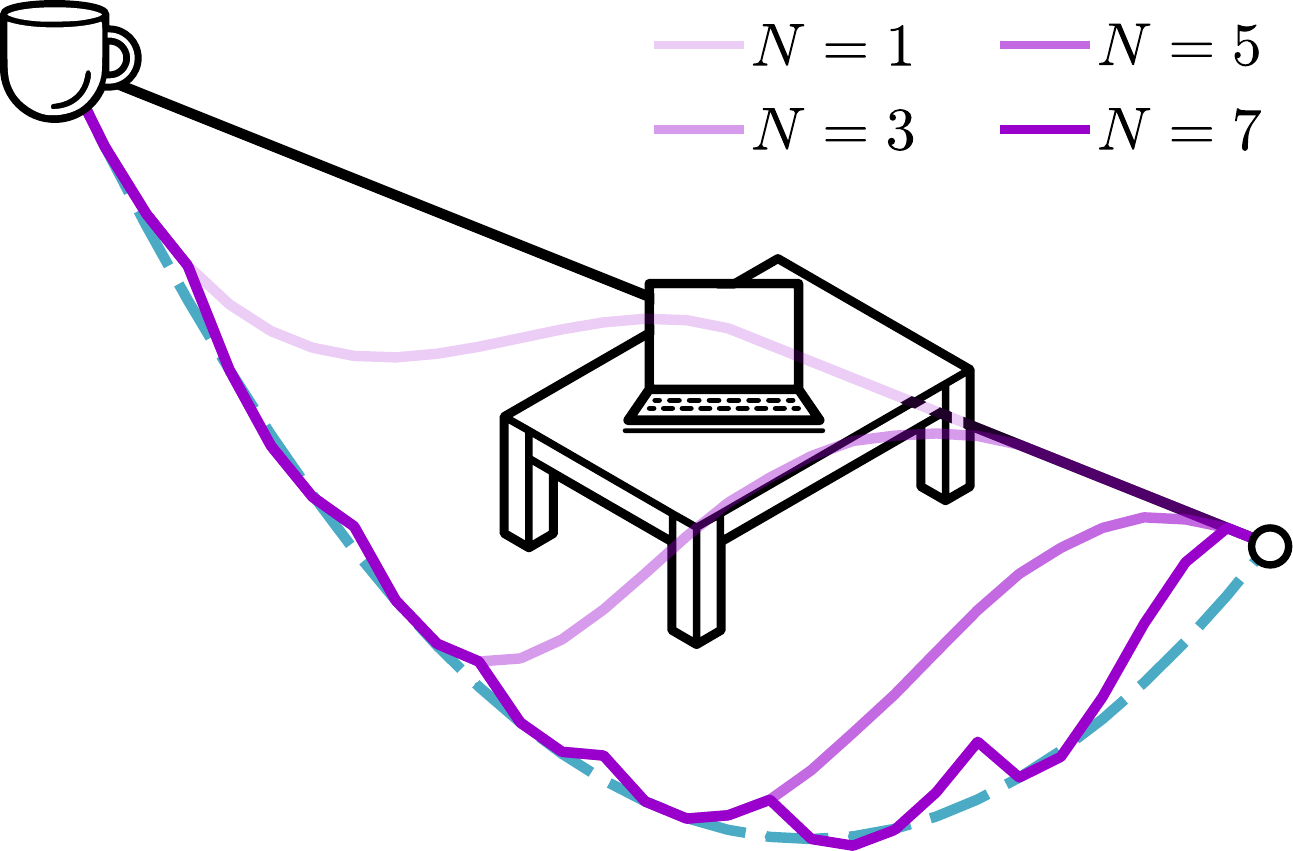}

		\caption{Responding to physical interaction by deforming the robot's trajectory. We propagate the human's interaction along the robot's trajectory to get $\xi_h$, the human's intended trajectory. We then set $\xi_h$ as the robot's trajectory. Here $N$ is the number of number of interactions: we show the robot's trajectory after $1$, $3$, $5$, and $7$ deformations. Importantly, when using deformations the robot never learns about task, but only updates its trajectory in the direction of the human's applied force.}

		\label{fig:sim2_deform}
	\end{center}

	\vspace{-0.5em}

\end{figure}

\begin{figure}[t]

	\begin{center}
		\includegraphics[width=0.8\columnwidth]{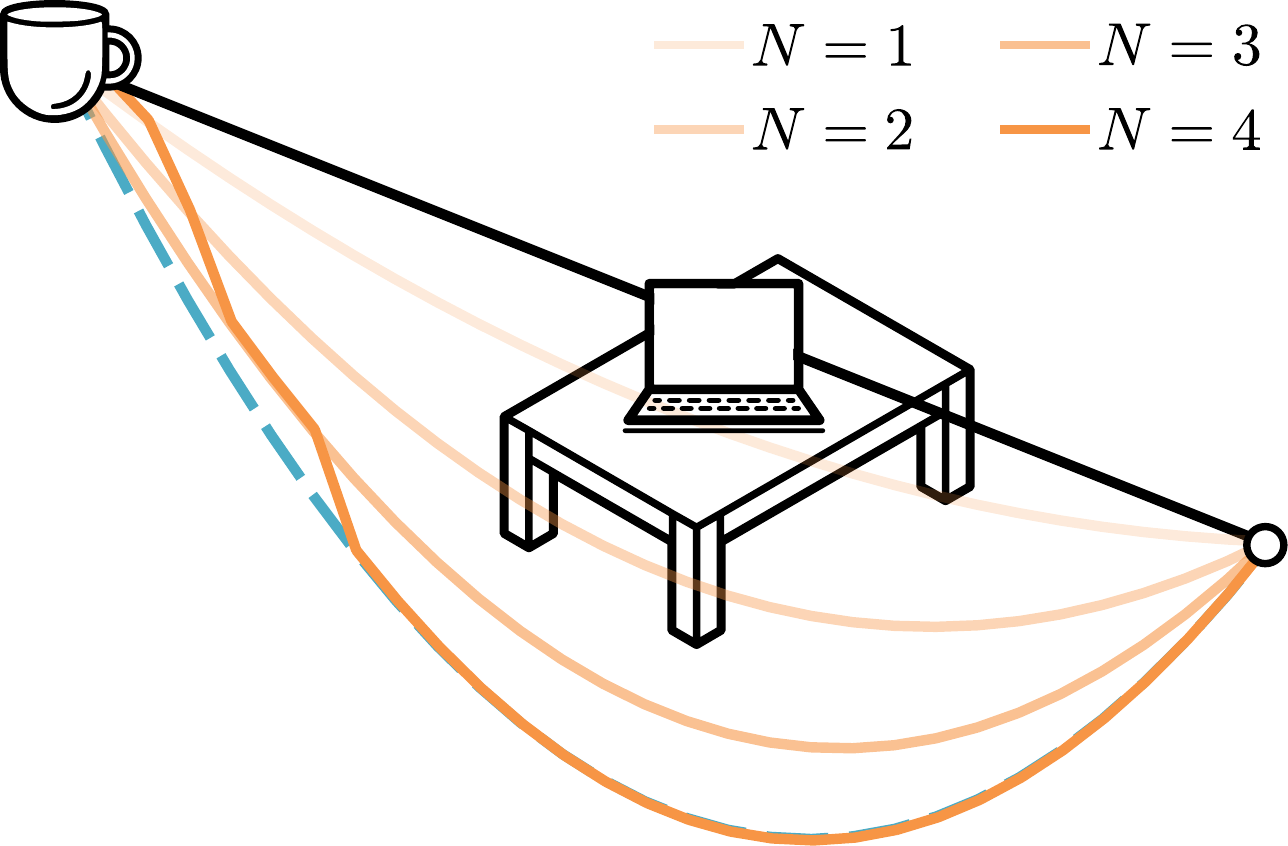}

		\caption{Responding to physical interactions using our proposed learning approach. As before, we propagate the human's interaction along the robot's current trajectory to get $\xi_h$, the human's intended trajectory. But now we go one step further: we compare $\xi_h$ to $\xi_r$ to update our estimate of the human's objective $\theta$. The robot then moves in the direction of the optimal trajectory for $\theta$. Under this approach the robot learns to avoid the laptop after $N=4$ corrections, and autonomously tracks the human's preferred trajectory (blue, dashed).}

		\label{fig:sim2_learn}
	\end{center}

	\vspace{-0.5em}

\end{figure}

\noindent \textbf{Learning vs. Deforming}. As part of our approximations we assumed that the human's interaction implies an \emph{intended trajectory}. Here we want to see whether learning from the intended trajectory---as in Algorithm~\ref{alg:a1}---is more optimal than simply setting that intended trajectory as the robot's trajectory. We compare two real-time learning methods: our \emph{Learning} approach, and the trajectory deformation method from \cite{losey2018trajectory}, which we refer to as \emph{Deforming}. The task used in these simulations is shown in Figs~\ref{fig:sim2_deform} and \ref{fig:sim2_learn}. Again, the robot is carrying a cup of coffee, but here the human would prefer for the robot to avoid carrying this coffee over their laptop. Thus, the two features are ``velocity'' and ``laptop.'' As before, the simulated human corrects the robot by guiding it back to their desired trajectory when the tracking error exceeds a predefined limit. In Deforming the robot does not learn about the human's objective, but instead propagates the human's corrections along the rest of the robot's trajectory. By contrast, in Learning we treat these trajectory deformations as the human's intended trajectory, which is then leveraged in our online update rule. Learning and Deforming can both be applied to change the robot's desired trajectory in real-time in response to pHRI, and Deforming is the same as treating the intended trajectory as the robot's trajectory.

\begin{figure}[t]

	\begin{center}
		\includegraphics[width=1\columnwidth]{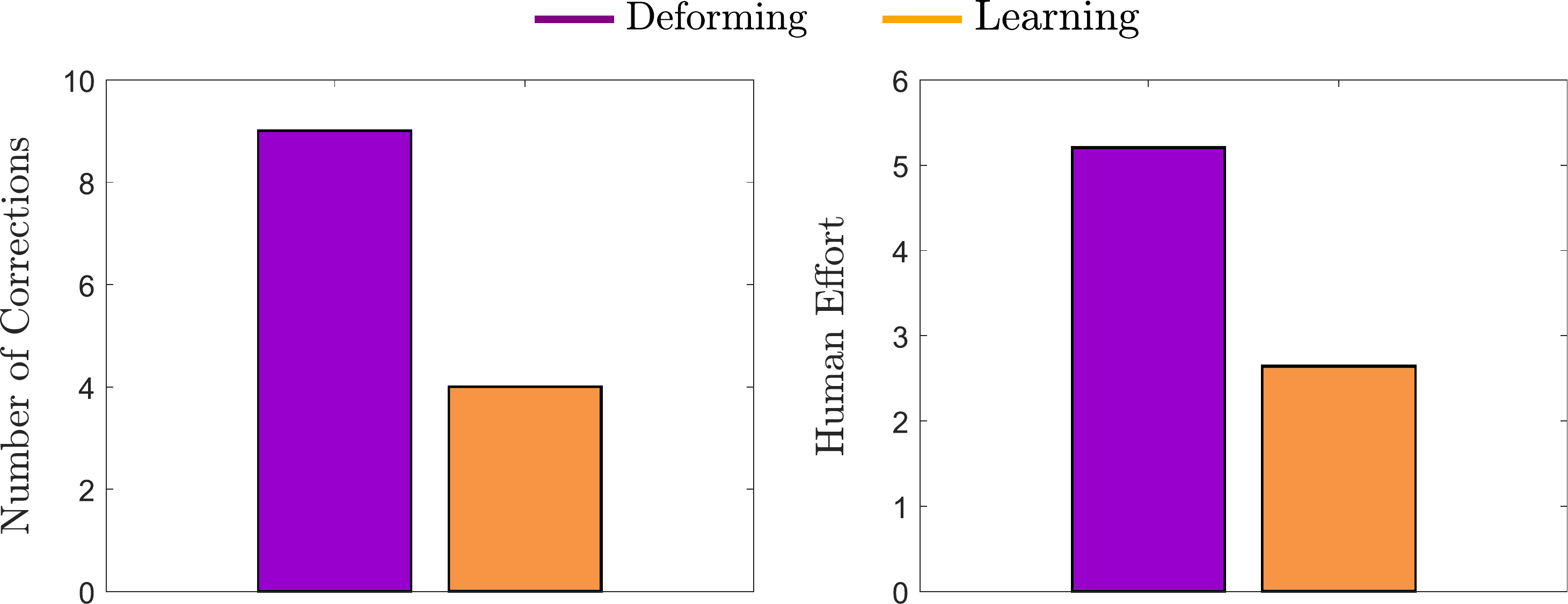}

		\caption{Comparison of Deforming and Learning across our simulations in Figs.~\ref{fig:sim2_deform} and \ref{fig:sim2_learn}. When robots only deform their trajectory in the direction of the human's applied force, humans must exert more effort and make more corrections to guide the robot's trajectory to their desired behavior. By contrast, Learning from these deformations enables the robot to correct not only the next few timesteps, but also to replan the remainder of the trajectory based on the human's correction.}

		\label{fig:sim2_metric}
	\end{center}

	\vspace{-1em}

\end{figure}

In Figs.~\ref{fig:sim2_deform} and \ref{fig:sim2_learn} we show the robot's trajectory after $N$ human corrections. Notice that Deformations result in local changes which aggregate over time, while---when we learn from these deformations---Learning replans the entire trajectory. Our findings are summarized in Fig~\ref{fig:sim2_metric}: it takes fewer corrections to track the human's desired trajectory with Learning, and the human also expends more effort with Learning. To make the comparison consistent, here we used the same propagation method from (\ref{eq:A6}) to get the Deformations and the intended trajectory for Learning. Based on our results, we conclude that Learning leads to more efficient online performance than Deformations alone, and, in particular, Learning requires \emph{less human effort} to complete the task correctly. 

\medskip

\begin{figure}[t]

	\begin{center}
		\includegraphics[width=0.8\columnwidth]{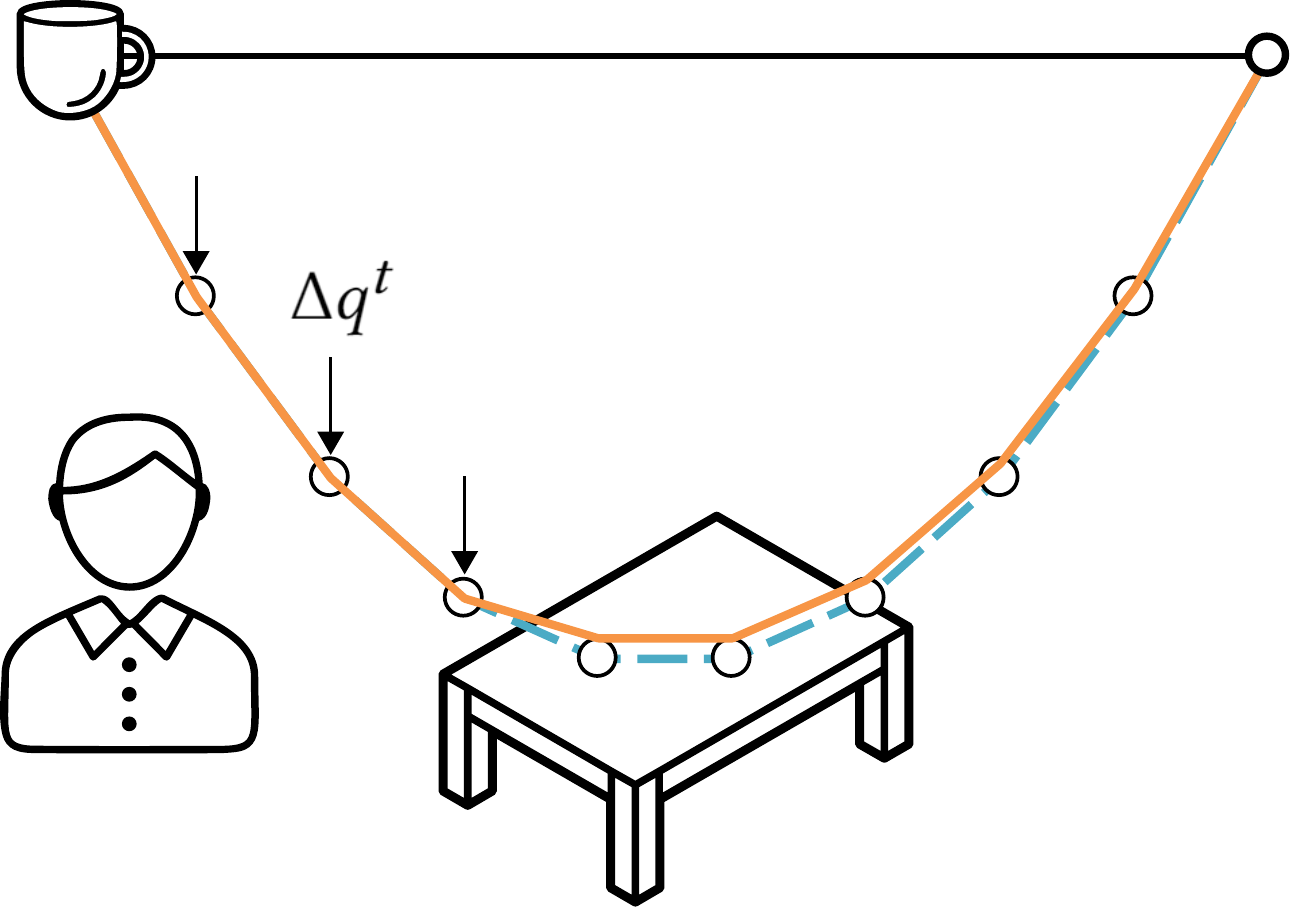}

		\caption{Comparing All-at-Once and One-at-a-Time learning with an \emph{optimal} simulated human. This human wants the robot to carry the coffee closer to table level, and provides physical corrections that exactly match their preferences. The human corrects the robot’s behavior over the first few timesteps (arrows) and the robot autonomously follows the human’s desired trajectory after these corrections. The robot's behavior is the same for All-at-Once and One-at-a-Time learning.}

		\label{fig:sim3_optimal}
	\end{center}

	\vspace{-0.5em}

\end{figure}

\begin{figure}[t]

	\begin{center}
		\includegraphics[width=0.8\columnwidth]{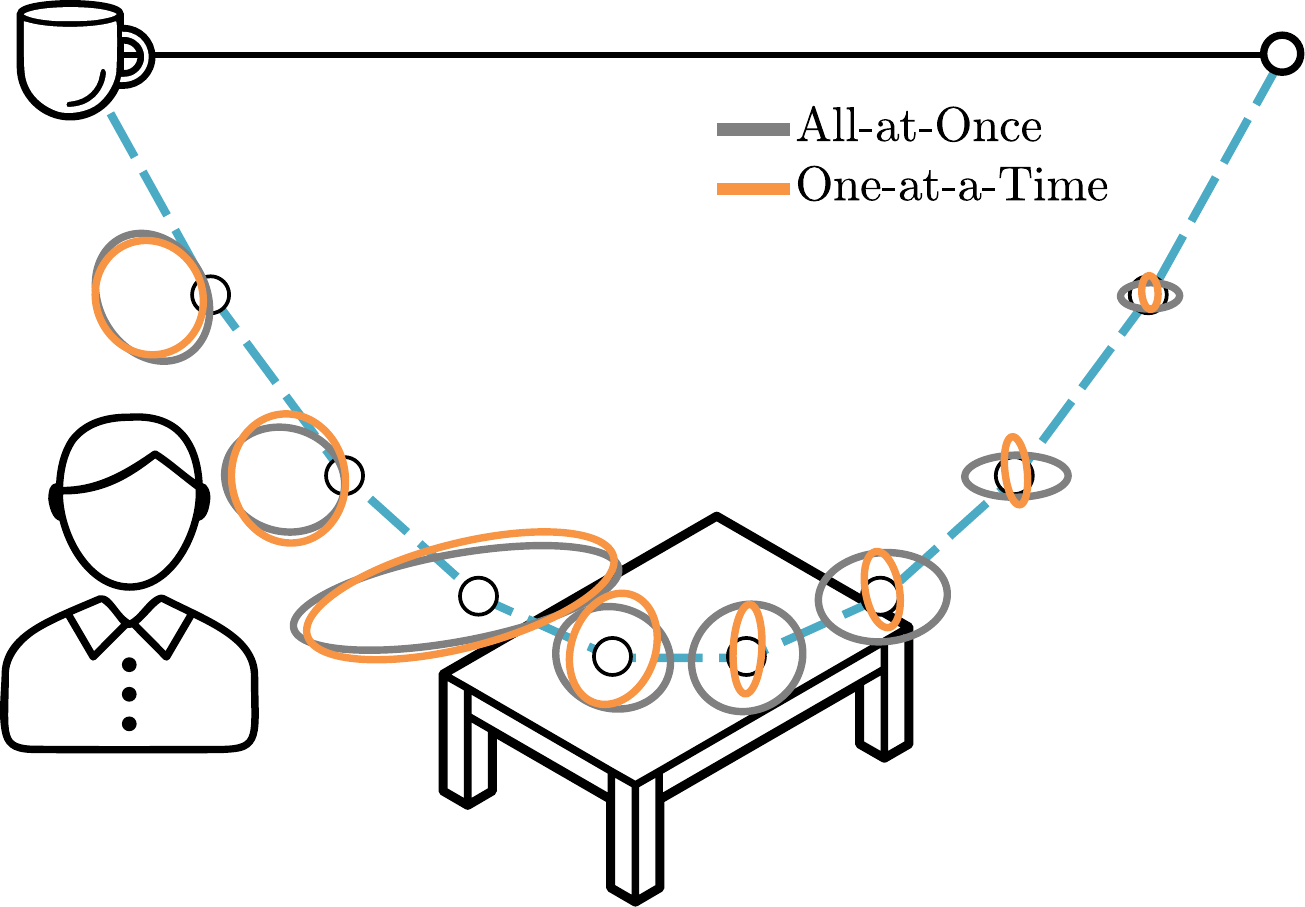}

		\caption{Comparing All-at-Once and One-at-a-Time learning with a \emph{noisy} simulated human. This noisy human wants the robot to move closer to the table, but accidentally provides biased corrections that also move the cup closer to the human. Ellipses show the robot's position at each timestep with $95\%$ confidence over $100$ simulations. The human unintentionally pulls the robot closer to their body at the start of the task, and with the All-at-Once approach they struggle to undo these mistakes in the second half of the task.}

		\label{fig:sim3_noisy}
	\end{center}

	\vspace{-0.5em}

\end{figure}

\noindent \textbf{All-at-Once vs. One-at-a-Time}. Previously we simulated tasks with only two features, and so a single feature weight was sufficient to capture the human's preference ($\theta \in \mathbb{R}$). In other words, either the all-at-once update or the one-at-a-time update could have been used for Learning. Now we compare \emph{All-at-Once} (\ref{eq:L5}) and \emph{One-at-a-Time} (\ref{eq:O5}) learning in a task with three features ($\theta \in \mathbb{R}^2$). This task is illustrated in Figs.~\ref{fig:sim3_optimal} and \ref{fig:sim3_noisy}. The human end-user trades off between the length of the robot's trajectory (velocity), the coffee's height above the table (table), and the robot's distance from the person (human). Like before, the weight associated with ``velocity'' is fixed, and the human's true objective is ${\theta = [0.5,0]}$, where $0.5$ is the weight associated with table and $0$ is the weight associated with human. Initially the robot believes that $\theta^0 = [0, 0]$, and therefore the robot is unaware that it should move closer to the table.

\begin{figure}[t]

	\begin{center}
		\includegraphics[width=1\columnwidth]{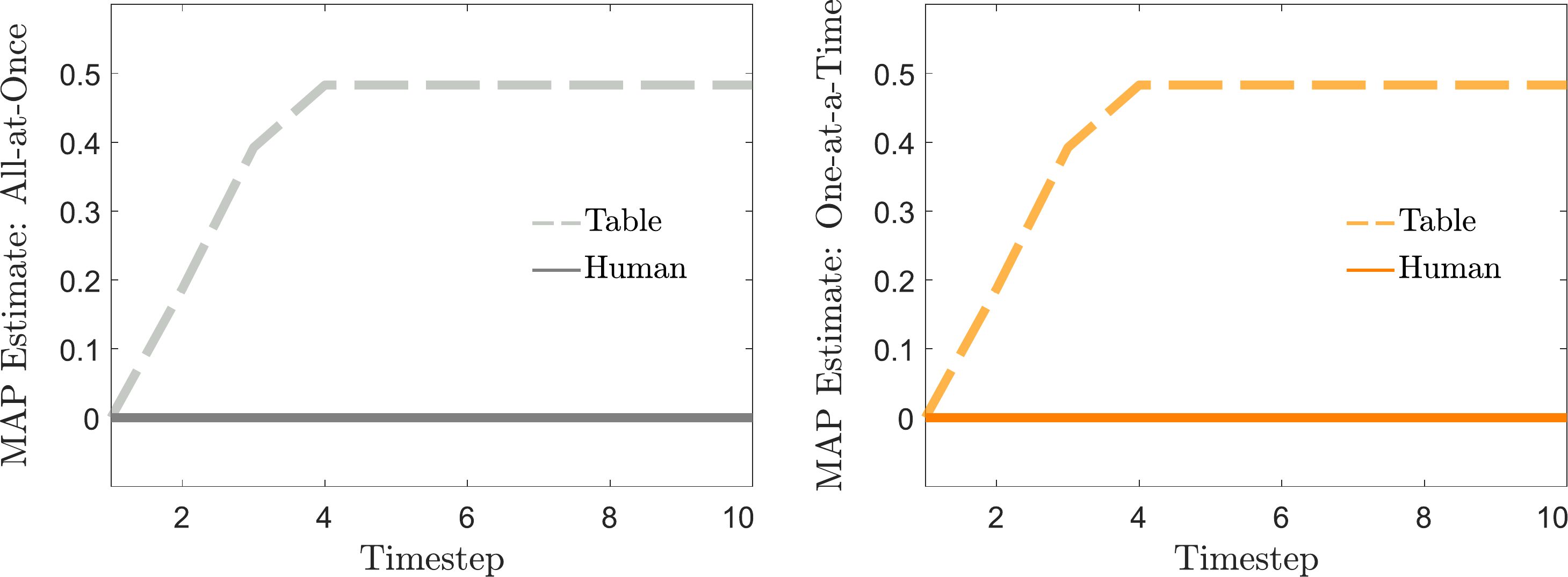}

		\caption{All-at-Once and One-at-a-Time learning with an \emph{optimal} simulated human. The true objective is $table = 0.5$, $human = 0$. Both All-at-Once and One-at-a-Time converge to the true objective: no unintentional corrections occur.}

		\label{fig:sim3_optimal_theta}
	\end{center}

	\vspace{-1em}

\end{figure}

We utilize two different simulated humans: (a) an \emph{optimal human}, who exactly guides the robot towards their desired trajectory, and (b) a \emph{noisy human}, who imperfectly corrects the robot's trajectory. Like in our previous simulations, the human intervenes to correct the robot when the robot's error with respect to their desired trajectory exceeds an acceptable margin of error: let us now refer to this as the optimal human. By contrast, the noisy human takes actions sampled from a Gaussian distribution which is centered at the optimal human's action. This distribution is biased in the direction of the human such that the noisy human tends to accidentally pull the robot closer to their body when correcting the table feature. Due to this noise and bias, the noisy human may \emph{unintentionally} correct the human feature.

\begin{figure}[t]

	\begin{center}
		\includegraphics[width=1\columnwidth]{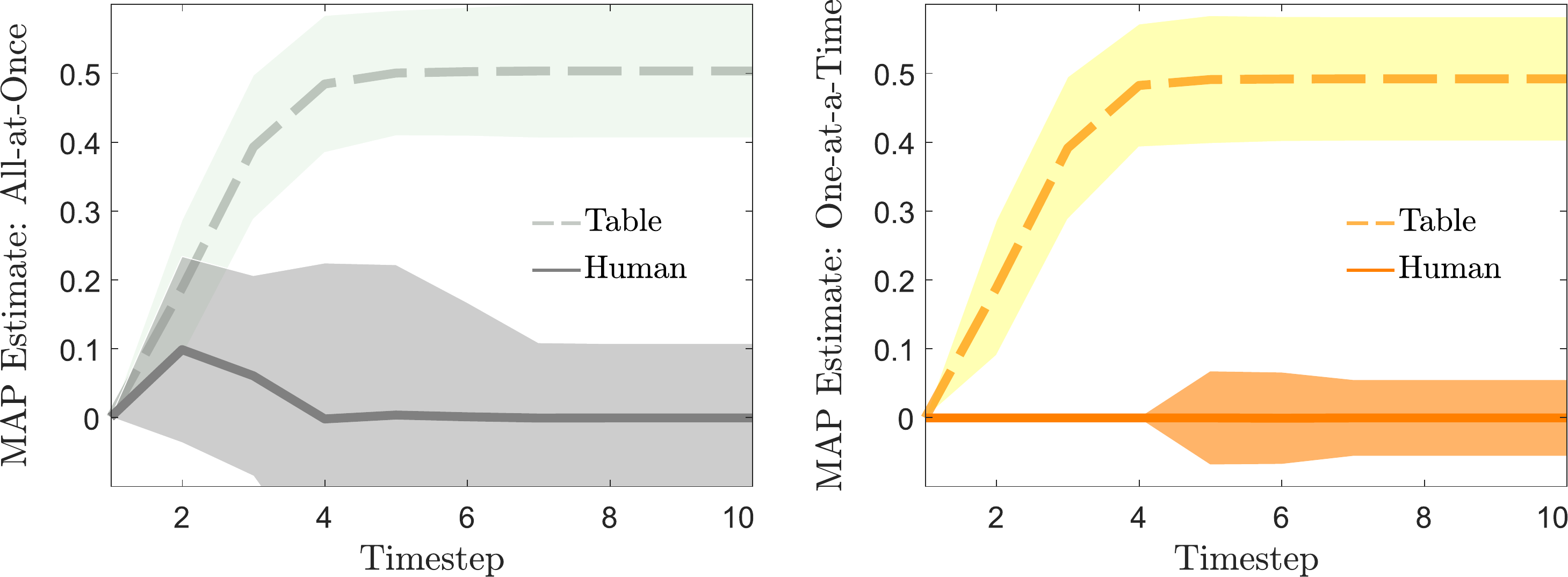}

		\caption{All-at-Once and One-at-a-Time learning with a \emph{noisy} simulated human. The shaded regions give the standard error of the mean. With All-at-Once, the robot initially learns that the human feature is important, and the person must undo that unintended learning. One-at-at-Time learning reduces the unintended effects of the human’s noisy corrections; the robot converges towards the human’s desired trajectory more rapidly.}

		\label{fig:sim3_noisy_theta}
	\end{center}

	\vspace{-1em}

\end{figure}

Our final simulation compares All-at-Once and One-at-a-Time learning for optimal and noisy humans. The results for an optimal human are shown in Figs.~\ref{fig:sim3_optimal} and \ref{fig:sim3_optimal_theta}, while the results for the noisy human are depicted in Figs.~\ref{fig:sim3_noisy} and \ref{fig:sim3_noisy_theta}. We find that the performance of All-at-Once and One-at-a-Time are identical when the human acts optimally: the robot accurately learns the importance of table, and does not change the weight of human. When the person acts noisily, however, One-at-a-Time learning causes better performance. More specifically, the noisy user corrected the All-at-Once robot during an average of $5.24$ timesteps, but only corrected the One-at-a-Time robot $3.56$ timesteps. Inspecting Fig.~\ref{fig:sim3_noisy_theta}, we observe that the noisy human unintentionally taught the human feature at the beginning of the task, and had to exert additional effort undoing this mistake on All-at-Once robots. We conclude that there is a benefit to One-at-a-Time learning when the human behaves noisily, since updating only one feature per timestep \emph{mitigates accidental learning}.

\begin{figure*}[ht]
    \centering
    \subfigure[Task 1: Keep the cup upright  \label{fig:task1_corl}]{\includegraphics[width=.32\textwidth]{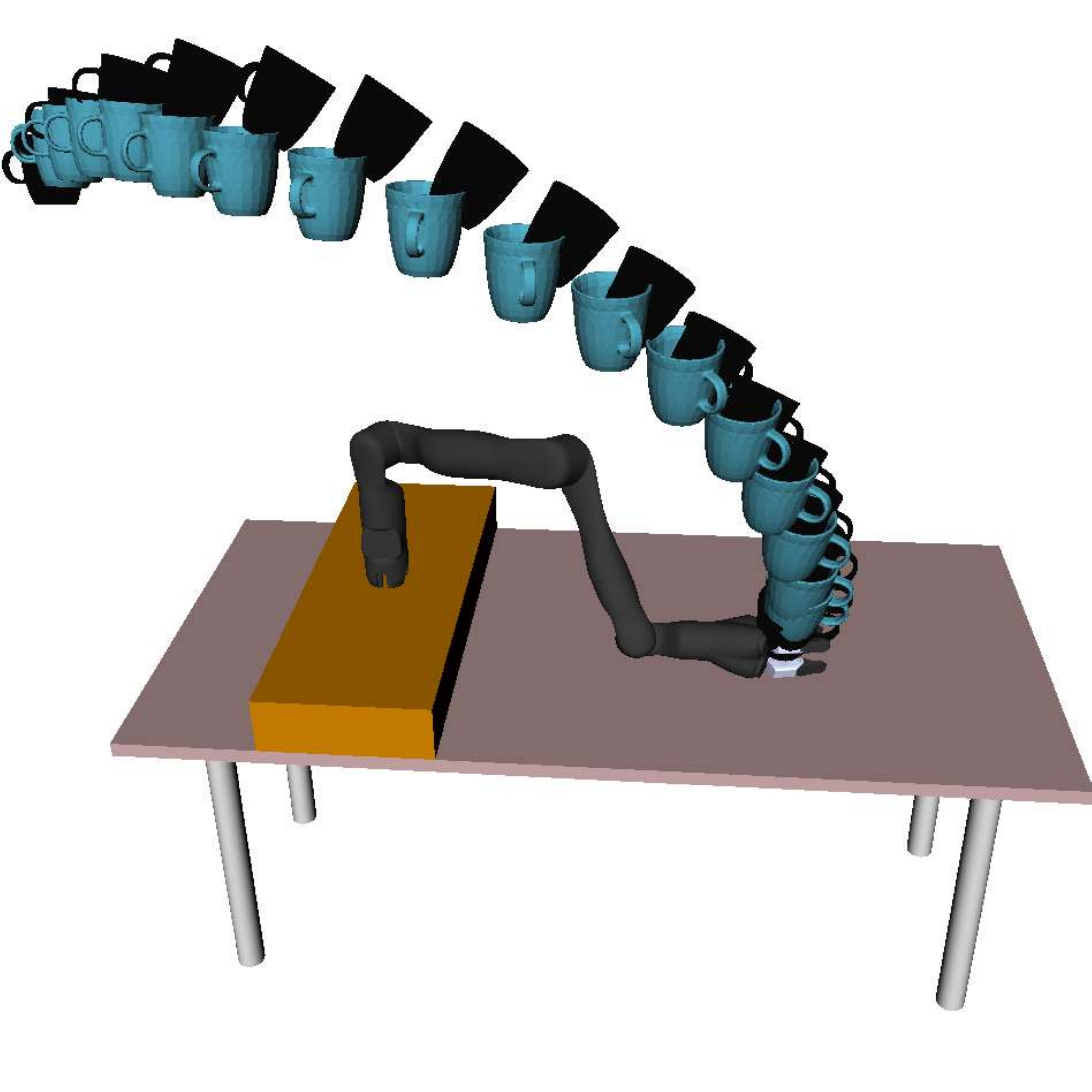}}\hfill
    \subfigure[Task 2: Carry closer to the table \label{fig:task2_corl}]{\includegraphics[width=.32\textwidth]{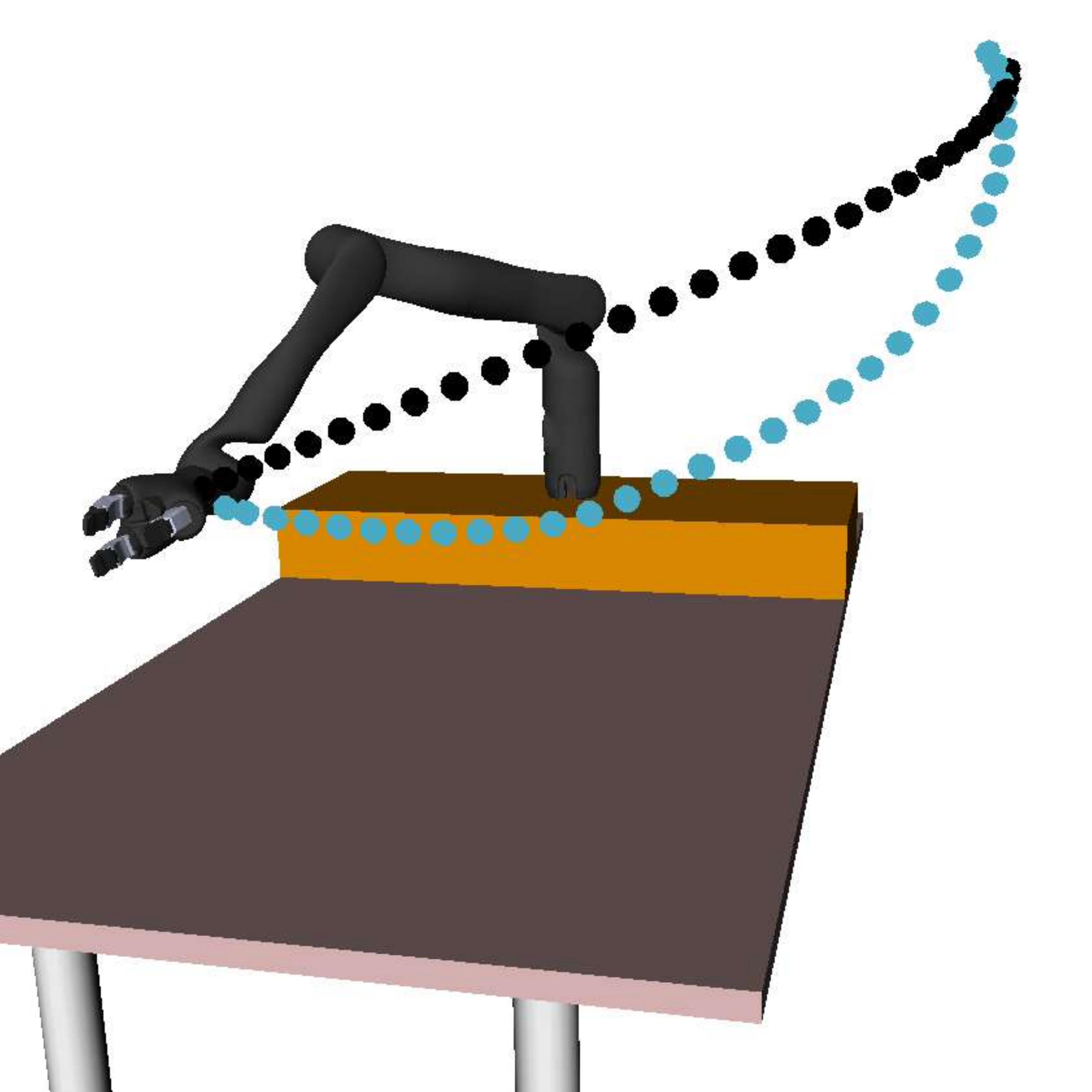}}\hfill
    \subfigure[Task 3: Avoid the region above a laptop \label{fig:task3_corl}]{\includegraphics[width=.32\textwidth]{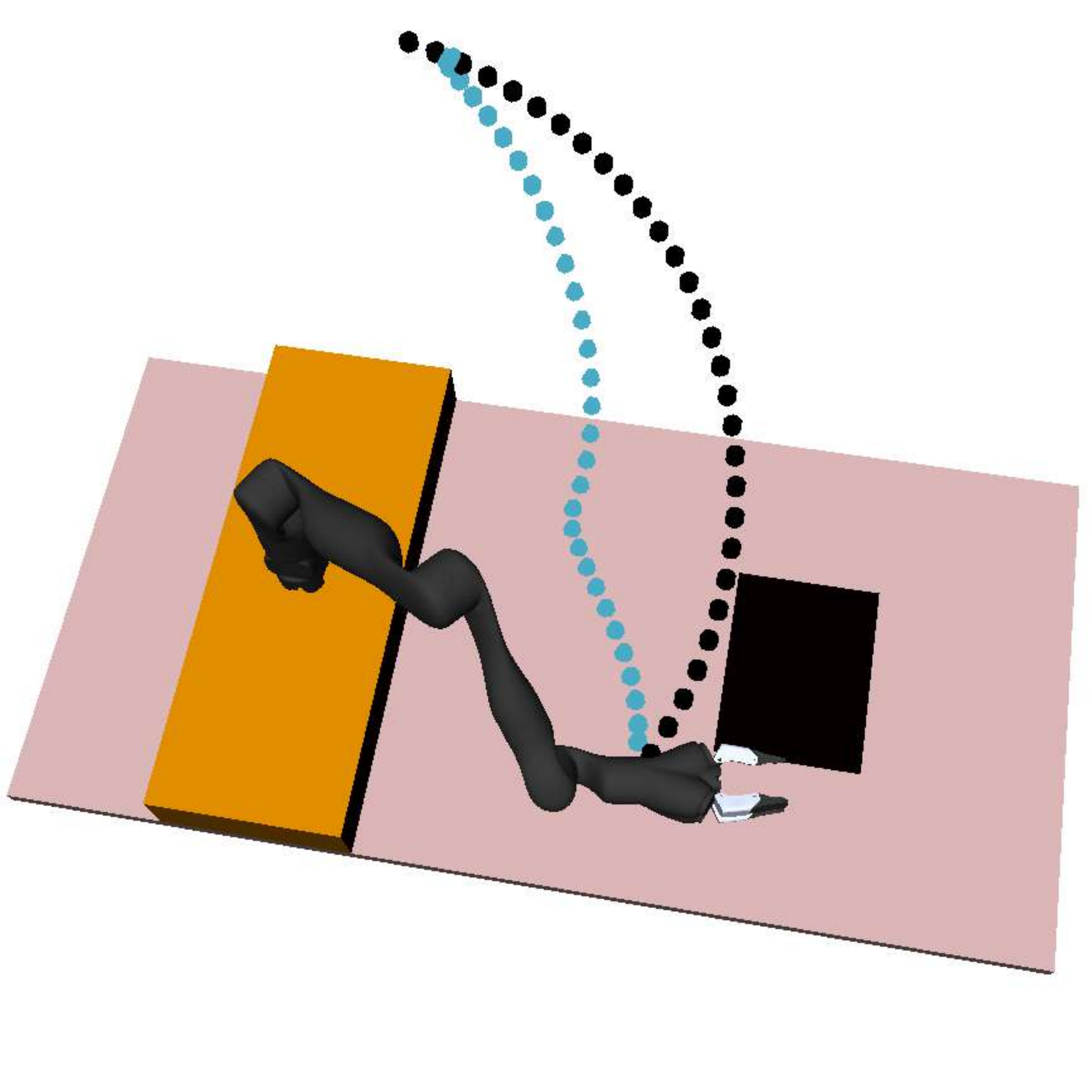}}\hfill
    \caption{Simulations depicting the robot trajectories for each of the three tasks in our first user study (Learning vs. Impedance). The black path represents the robot's initial trajectory, and the blue path represents the human's desired trajectory.} \label{fig:tasks_corl}
\end{figure*}

\section{User Studies}


To evaluate the benefits of using physical interaction to communicate we conducted two user studies with a 7-DoF robotic arm (JACO2, Kinova). In the first study, we tested \emph{whether} learning from pHRI is useful when humans interact, and compared our online learning approach to a state-of-the-art response that treated interactions as disturbances (Learning vs. Impedance). In the second study, we tested \emph{how} the robot should learn from end-users, and compared one-at-a-time learning to all-at-once learning (One-at-a-Time vs. All-at-Once). During both studies the participants and the robot worked in close physical proximity. In all experimental tasks, the robot began with the wrong objective function, and participants were instructed to physically interact with the robot to correct its behavior\footnote{For video footage of the experiment, see: \url{https://youtu.be/I2YHT3giwcY}}.

\begin{figure}[t!]
\centering
\includegraphics[width=1\columnwidth]{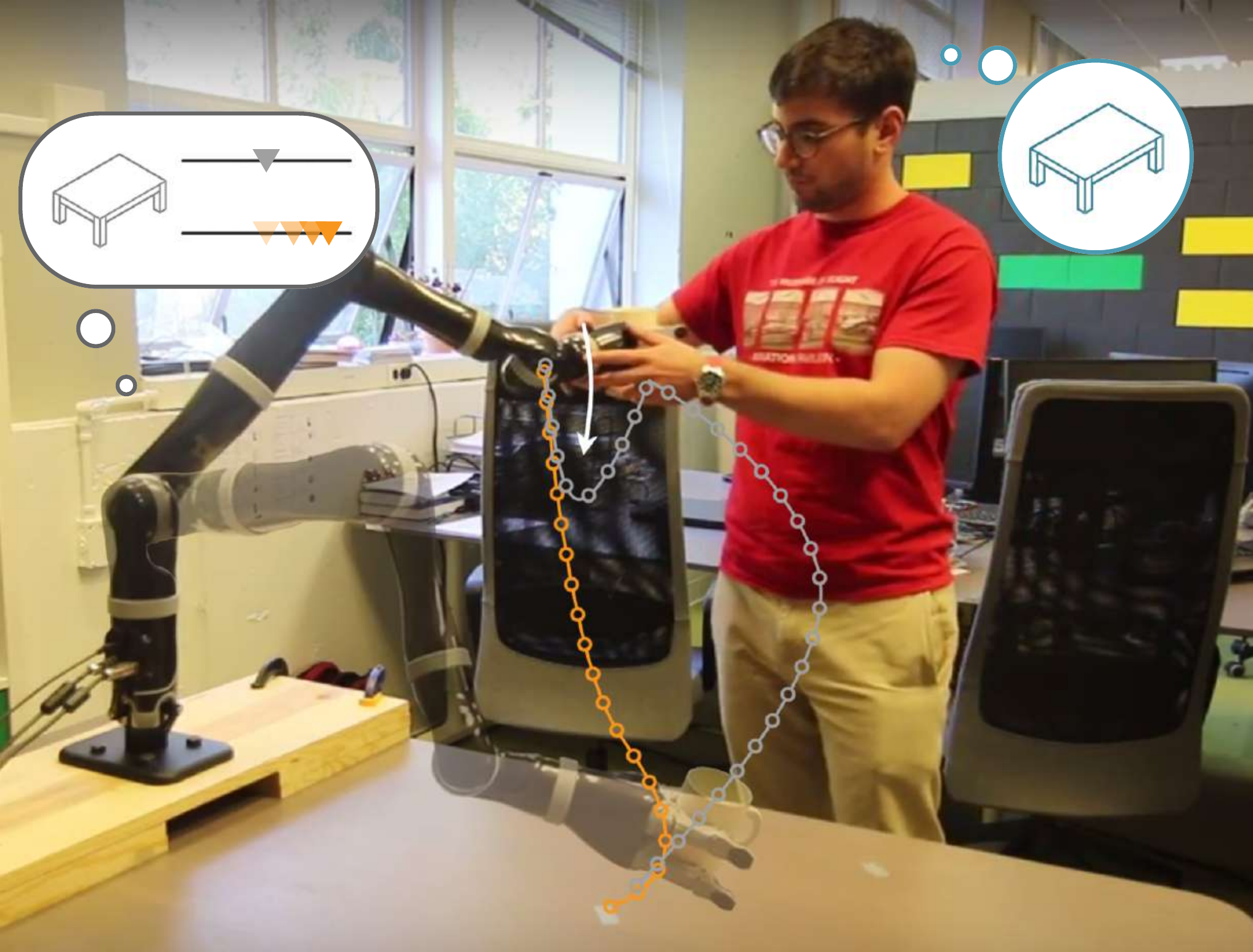}
\caption{During the first user study participants interacted with a robot that maintained a fixed objective (Impedance, grey) and a robot that learned from their physical interactions to update its objective (Learning, orange).} \label{fig:corlfig}
\end{figure}

\subsection{Learning vs. Impedance}

We have argued that pHRI is a means for humans to correct the robot's behavior. In our first user study, we compare a robot that treats human interactions as intentional (and learns from them) to a robot that assumes all human interactions are disturbances (and ignores them).

\smallskip


\begin{figure*}[h]
\begin{multicols}{2}
\centering
\includegraphics[width=.45\textwidth]{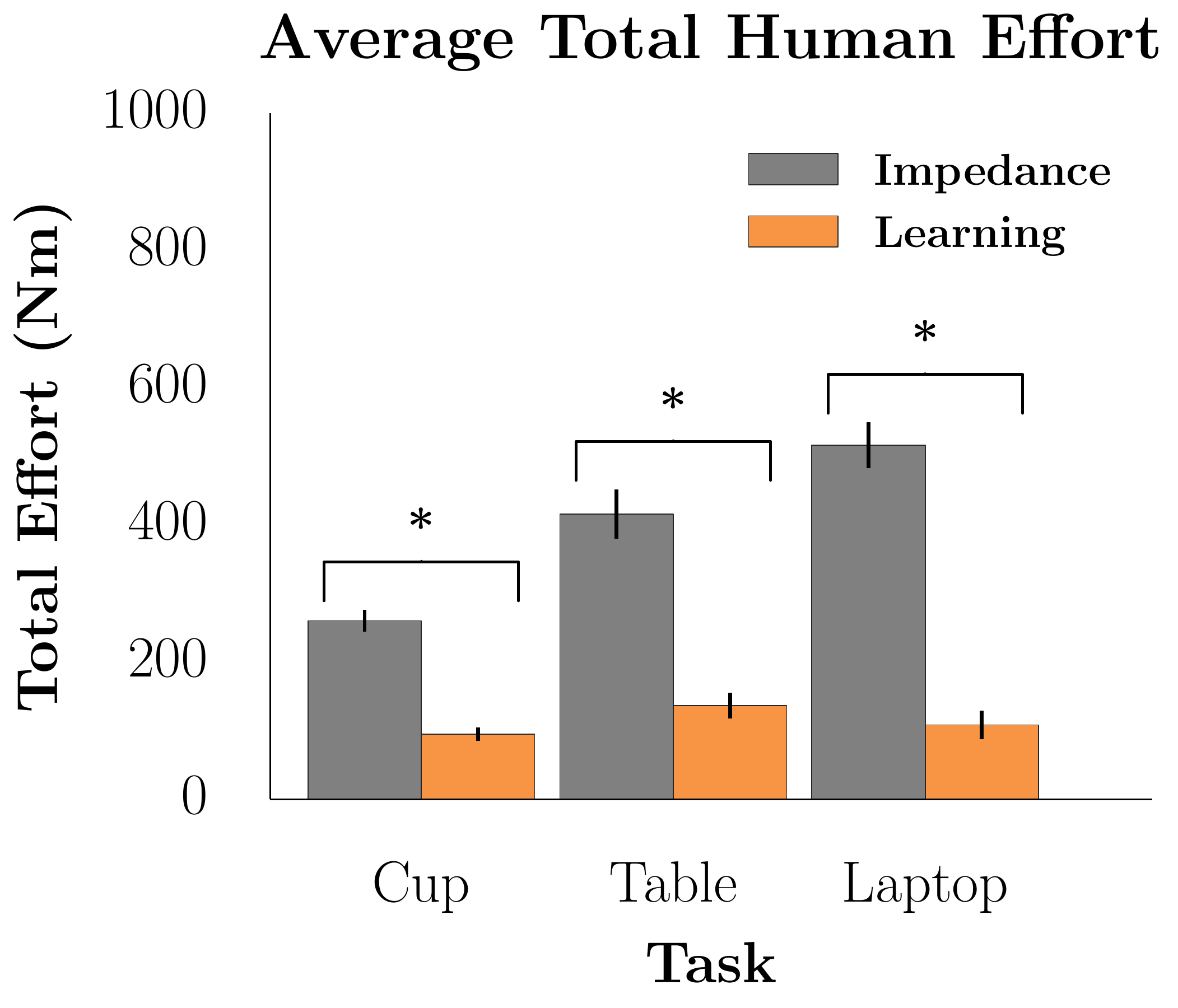}\\
\includegraphics[width=.45\textwidth]{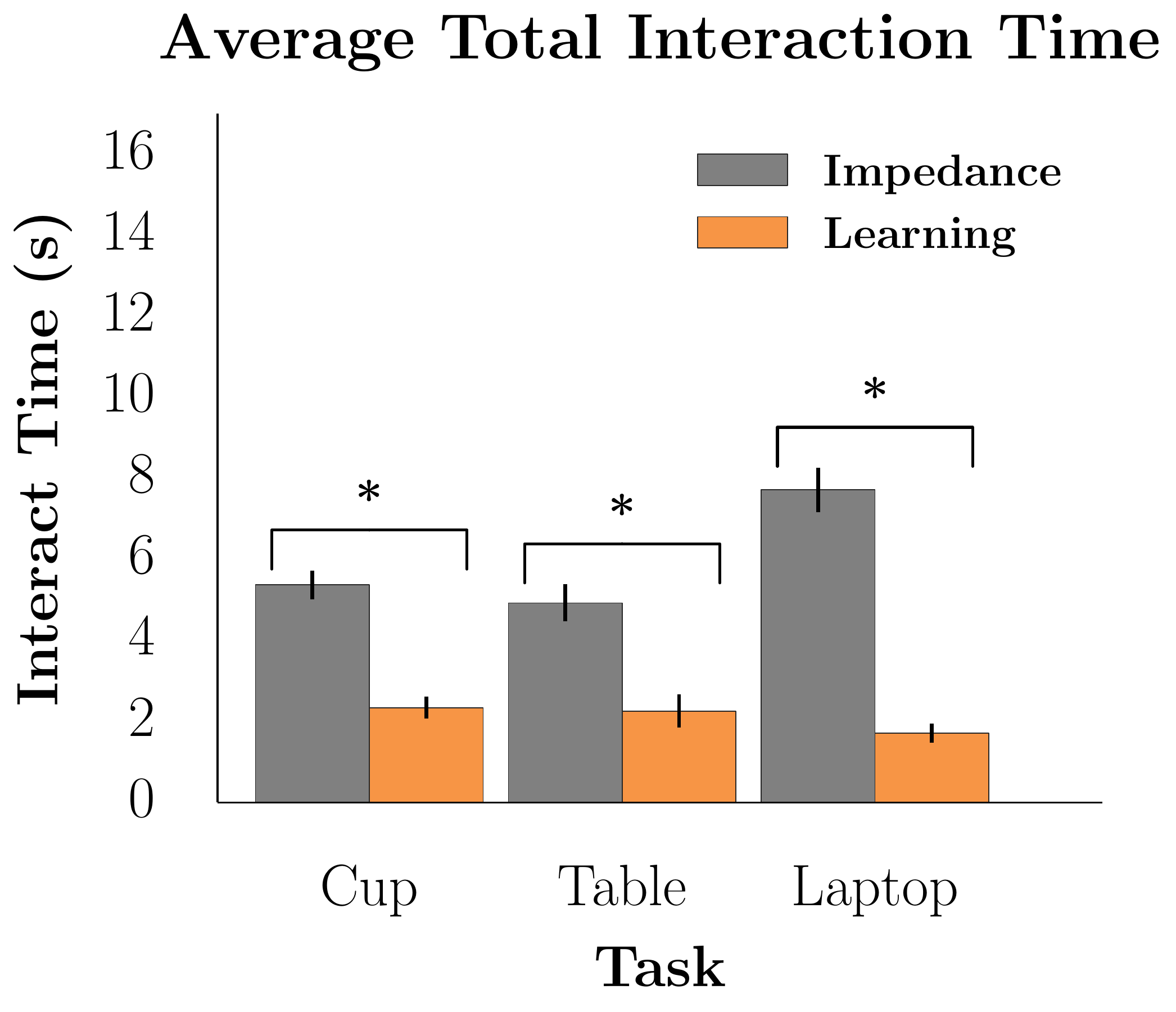}\\
\end{multicols}
\vspace{-2em}
\caption{Objective results from our first user study. We explored whether robots should learn from physical interactions (Learning vs Impedance). Learning from pHRI decreased participant effort and interaction time across all experimental tasks (the total trajectory time was 15s). An asterisk (*) means $p < .0001$.
}
\label{fig:iact_results}
\end{figure*}

\begin{figure*}[ht!]
\begin{multicols}{2}
\centering
\includegraphics[height=.4\textwidth]{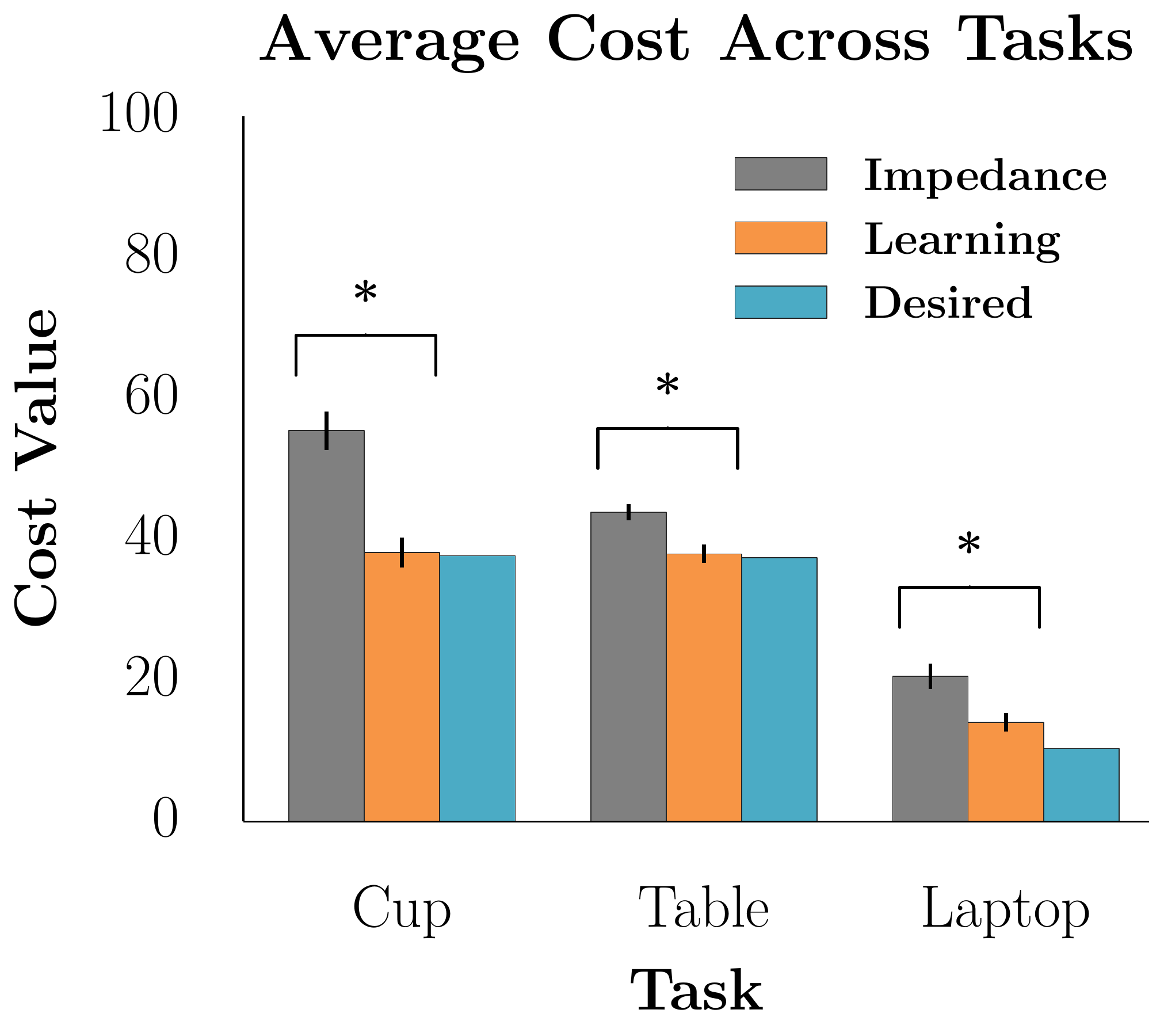}\\
\includegraphics[height=.35\textwidth]{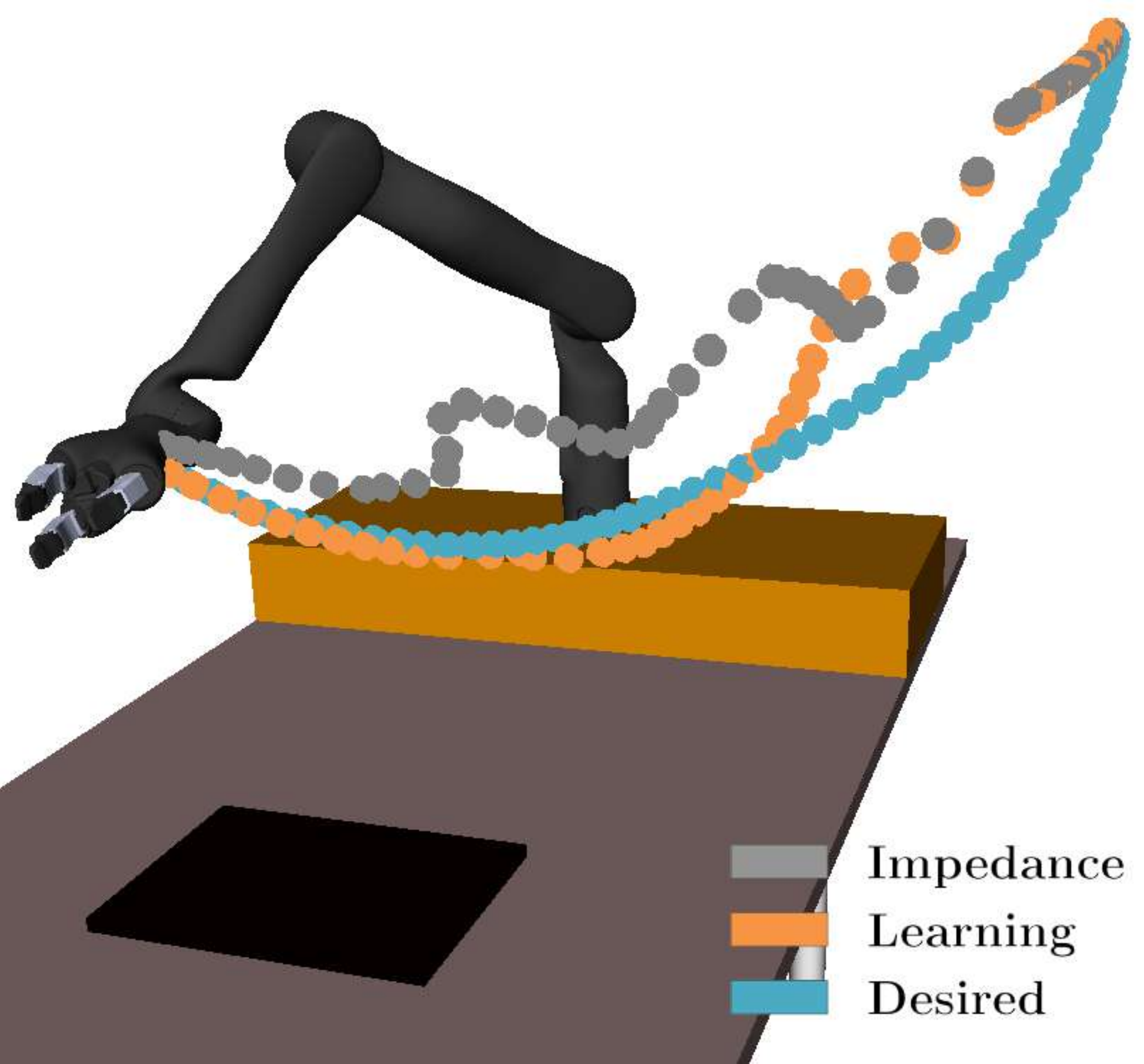}\\
\end{multicols}
\vspace{-2em}
\caption{(Left) Average cost for each task and the cost of the desired trajectory. Robots that always follow the human's desired trajectory minimize cost. An asterisk (*) means $p < 0.0001$. (Right) Plot of sample participant data from the laptop task: the desired trajectory is in blue, the trajectory with the Impedance condition is in gray, and the Learning condition trajectory is in orange.}
\label{fig:traj_results}
\end{figure*}

\noindent\textbf{Independent Variables.} We manipulated the \textit{pHRI strategy} with two levels: \emph{Learning} and \emph{Impedance}. The Learning robot used our proposed method (Algorithm \ref{alg:a1}) to react to physical corrections and re-plan a new trajectory during the task. By contrast, the Impedance robot used impedance control (our method without updating $\hat{\theta}$) to react to physical interactions and then return to the originally planned trajectory. Because impedance control is currently the most common strategy for responding to pHRI \citep{haddadin2016physical}, we treated Impedance as the state-of-the-art.

\smallskip

\noindent\textbf{Dependent Measures.} We measured the robot's objective performance with respect to the human's actual objective. One challenge in designing our experiment was that each participant might have a different internal objective $\theta$ for any given task depending on their experiences and preferences. Since we did not have direct access to every person's internal preferences, we defined the true objective $\theta$ ourselves, and conveyed the objective to participants by demonstrating the desired optimal robot behavior. We instructed participants to correct the robot to achieve this behavior with as little interaction as possible. To understand how users perceived the robot, we also asked subjects to complete a 7-point Likert scale survey for both pHRI strategies: the questions from this survey are shown in Table~\ref{tab:likert}.

\smallskip

\noindent\textbf{Hypotheses.}
\begin{displayquote}
\textbf{H1.} \emph{Learning will decrease interaction time, effort, and cumulative trajectory cost.}
\end{displayquote}
 
\begin{displayquote}
\textbf{H2.} \emph{Learning users will believe the robot understood their preferences, feel that interacting with the robot was easier, and perceive the robot as more predictable and collaborative.}
\end{displayquote}

\smallskip

\noindent\textbf{Tasks.} We designed three household manipulation tasks for the robot to perform in a shared workspace, in addition to one familiarization task. The robot's objective function consisted of two features: ``velocity'' and a task-specific feature, where $\Phi(\xi) \in [0,1]$. Because one feature weight was sufficient to capture these tasks (i.e., $\theta \in \mathbb{R}$) both the all-at-once and one-at-a-time learning approached were here identical. For each task, the robot carried a cup from a start pose to a goal pose with an \textit{initially incorrect objective}, forcing participants to correct its behavior during the task. 

In the familiarization task the robot's original trajectory moved too close to the human. Participants had to physically interact with the robot to make the robot keep the cup farther away from their body. In Task 1 the robot carried a cup directly from start to goal, but did not realize that it needed to keep this cup upright. Participants had to intervene to prevent the cup from spilling. In Task 2 the robot carried the cup too high in the air, risking breaking that cup if it were to slip. Participants had to correct the robot to keep the cup closer to the table. Finally, in Task 3 the robot moved the cup over a laptop to reach its final goal pose, and participants physically guided the robot away from this laptop region. We include a depiction of our three experimental tasks in Fig.~\ref{fig:tasks_corl}.

\medskip

\noindent\textbf{Participants.} We employed a within-subjects design and counterbalanced the order of the pHRI strategy conditions. Ten total members of the UC Berkeley community ($5$ male, $5$ female, age range $18$-$34$) provided informed consent according to the approved IRB protocol and participated in the study. All participants had technical backgrounds. None of the participants had prior experience interacting with the robot used in our experiments. 


\medskip

{\renewcommand{\arraystretch}{1.5}
\begin{table*}[t]
 \centering
 \adjustbox{max width=\textwidth}{
 \begin{tabular}{|c|l|c|c|c|c|c|}
 \hline
 \rowcolor{Gray}
 & \multicolumn{1}{c|}{\textbf{Questions}} & \multicolumn{1}{c|}{\textbf{Cronbach's $\alpha$}} & \multicolumn{1}{c|}{\textbf{Imped LSM}} & \multicolumn{1}{c|}{\textbf{Learn LSM}} & \multicolumn{1}{c|}{\textbf{F(1,9)}} & \multicolumn{1}{c|}{\textbf{p-value}} \\
 \hline
 \parbox[t]{2mm}{\multirow{4}{*}{\rotatebox[origin=c]{90}{\textbf{understanding}}}} 
 & By the end, the robot understood how I wanted it to do the task. & \multirow{4}{*}{0.94} & \multirow{4}{*}{1.70} & \multirow{4}{*}{5.10} & \multirow{4}{*}{118.56} & \multirow{4}{*}{\textbf{$<$.0001}}\\
 & Even by the end, the robot still did not know how I wanted it to do the task. &&&&&\\
 & The robot learned from my corrections. &&&&&\\
 & The robot did not understand what I was trying to accomplish.  &&&&&\\
 \hline
 \parbox[t]{2mm}{\multirow{2}{*}{\rotatebox[origin=c]{90}{\textbf{effort}}}} 
 & I had to keep correcting the robot. & \multirow{2}{*}{0.98} & \multirow{2}{*}{1.25} & \multirow{2}{*}{5.10} & \multirow{2}{*}{85.25} & \multirow{2}{*}{\textbf{$<$.0001}}\\
 & The robot required minimal correction. & &&&&\\
 \hline
 \parbox[t]{2mm}{\multirow{2}{*}{\rotatebox[origin=c]{90}{\textbf{predict}}}} 
 & It was easy to anticipate how the robot will respond to my corrections. & 0.8 & 4.90 & 4.70 & 0.06 & 0.82\\
 & The robot's response to my corrections was surprising.  & 0.8 & 3.10 & 3.70 & 0.89 & 0.37\\
 \hline
 \parbox[t]{2mm}{\multirow{2}{*}{\rotatebox[origin=c]{90}{\textbf{collab}}}} 
 & The robot worked with me to complete the task.  & \multirow{2}{*}{0.98} & \multirow{2}{*}{1.80 } & \multirow{2}{*}{4.80} & \multirow{2}{*}{55.86} & \multirow{2}{*}{\textbf{$<$.0001}}\\
 & The robot did not collaborate with me to complete the task. &&&&&\\
 \hline
 \end{tabular}
 }
 \strut 
 \caption{Subjective ratings collected from a 7-point Likert scale survey. Participants answered each question once after working with the Impedance condition, and once after the Learning condition. The four question scales are shown on the left. Imped is short for Impedance, Learn is short for Learning, and LSM stands for Likert scale mean. Higher LSM values are better (more understanding, less effort, more predictable, more collaborative). ANOVA results are on the far right.} \label{tab:likert}
 \end{table*}}

\noindent\textbf{Procedure.} For each pHRI strategy participants performed the familiarization task, followed by the three experimental tasks, and then filled out our user survey. They attempted every task twice during each pHRI strategy for robustness (we recorded the attempt number for our analysis). Since we artificially set the true objective $\theta$, we showed participants both the original and desired robot trajectory before the task started to make sure that they understood this objective and got a sense of the corrections they would need to make. 

\medskip

\noindent\textbf{Results -- Objective.} We conducted a repeated measures ANOVA with pHRI strategy (Impedance or Learning) and trial number (first attempt or second attempt) as factors. We applied this ANOVA to three objective metrics: total participant effort, interaction time, and cost\footnote{For simplicity, we only measured the value of the feature that needed to be modified in each task, and computed the absolute difference from the feature value of the optimal trajectory.}. Fig.~\ref{fig:iact_results} shows the results for human effort and interaction time, and Fig.~\ref{fig:traj_results} shows the results for cost. Learning resulted in significantly less interaction force ($F(1,116) = 86.29, p < 0.0001$) interaction time ($F(1,116) = 75.52, p < 0.0001$), and task cost ($F(1,116) = 21.85, p < 0.0001$). Interestingly, while trial number did not significantly affect participant's performance with either method, attempting the task a second time yielded a marginal improvement for the impedance strategy but not for the learning strategy. This may suggest that it is easier for users to familiarize themselves with the impedance strategy. 

Overall, our results support \textbf{H1}. Using interaction forces to learn about the objective $\theta$ here enabled the robot to better complete its tasks with less human effort when compared to a state-of-the-art impedance controller. 

\medskip

\noindent\textbf{Results -- Subjective.} Table \ref{tab:likert} shows the results of our participant survey. We tested the reliability of four scales, and found the understanding, effort, and collaboration scales to be reliable. Thus, we grouped each of these scales into a combined score, and ran a one-way repeated measures ANOVA on each resulting score. We found that the robot using our Learning method was perceived as significantly ($p < 0.0001$) more understanding, less difficult to interact with, and more collaborative than the Impedance approach.

By contrast, we found no significant difference between our Learning method and the baseline Impedance method in terms of predictability. Participant comments suggest that while the robot quickly adapted to their corrections when Learning  (e.g. ``the robot seemed to quickly figure out what I cared about and kept doing it on its own"), determining what the robot was doing during Learning was less intuitive (e.g. ``if I pushed it hard enough sometimes it would seem to fall into another mode, and then do things correctly").

We conclude that \textbf{H2} was partially supported: although users did not perceive Learning to be more predictable than Impedance, participants believed that the Learning robot understood their preferences better, took less effort to interact with, and was a more collaborative partner. 

\medskip

\noindent\textbf{Summary.} Robots that treat pHRI as a source of information (rather than as a disturbance) are capable of online, in-task learning. Learning from pHRI resulted in better objective and subjective performance than a traditional Impedance approach. We found that the Learning robot better matched the human's preferred behavior with less human effort and interaction time, and participants perceived the Learning robot as easier to understand and collaborate with. However, participants did not think that the Learning robot was more predictable than the Impedance robot.

\begin{figure*}[ht]
    \centering
    \subfigure[Task 1: Correct one feature, the distance to table (table)  \label{fig:task1_hri}]{\includegraphics[width=.45\textwidth]{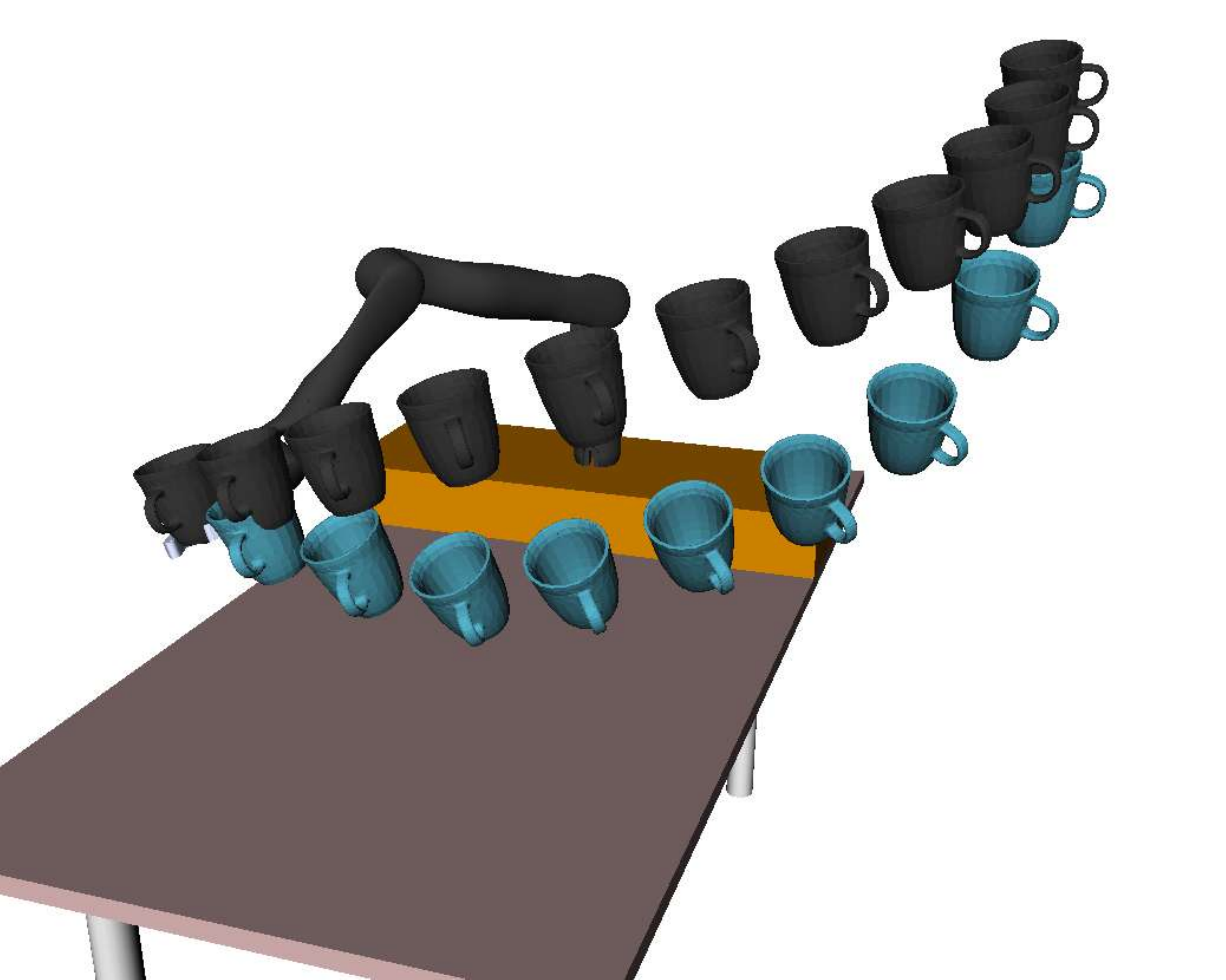}}\quad \quad
    \subfigure[Task 2: Correct two features, the cup orientation (cup) and the distance to table (table) \label{fig:task2_hri}]{\includegraphics[width=.45\textwidth]{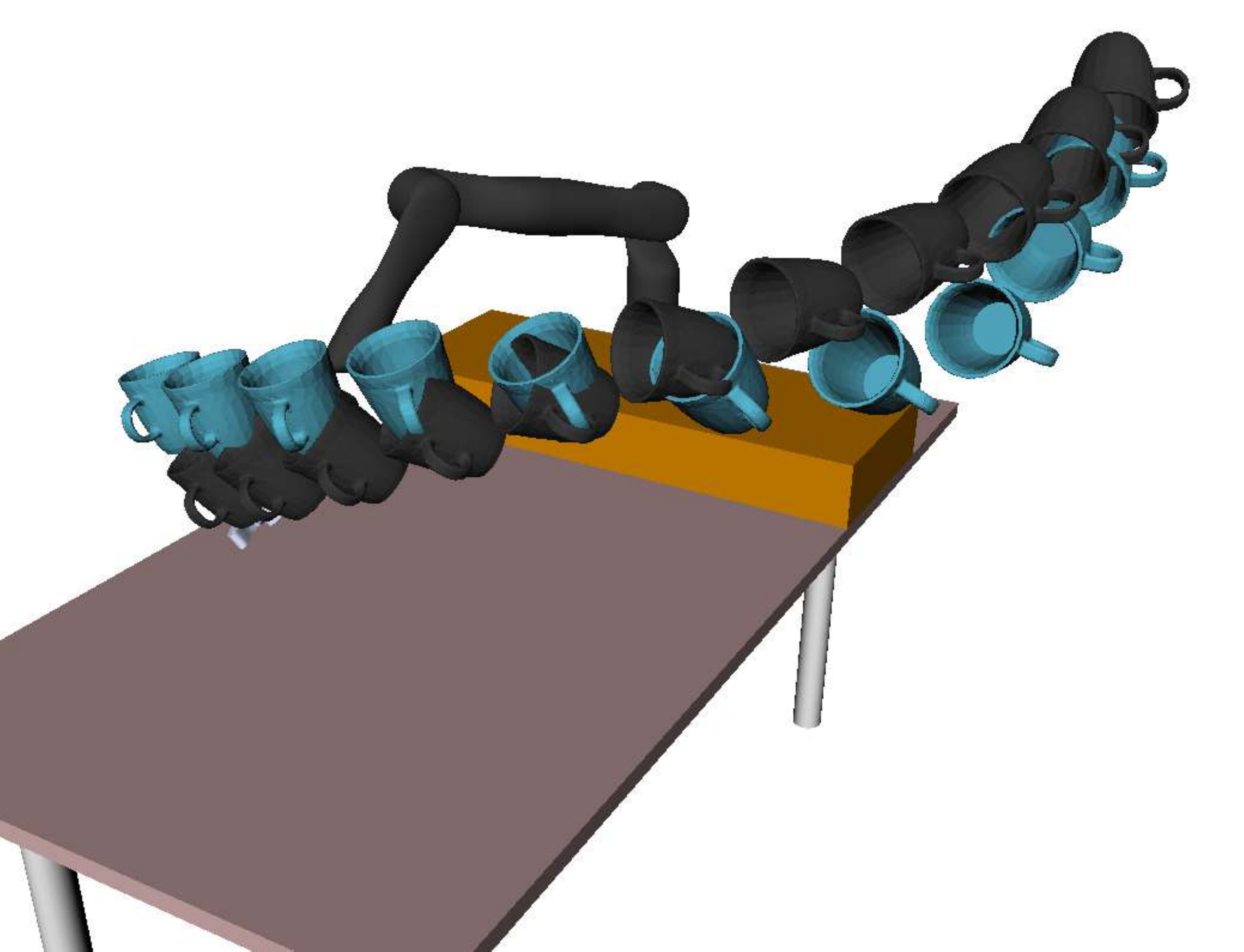}}
    \caption{Simulations depicting the robot trajectories for both of the two tasks in our second user study (One-at-a-Time vs. All-at-Once). The black path represents the robot's original trajectory, and the blue path represents the human's desired trajectory. Note that the robot now has multiple features, making it possible for the human to accidentally correct one or both features.} \label{fig:tasks_hri}
    \vspace{0.5em}
\end{figure*}

\subsection{One-at-a-Time vs. All-at-Once}

We have found that learning from pHRI is beneficial; now we want to determine \emph{how} the robot should learn. In our second user study we focused on objective functions which encode multiple task-related features. In these scenarios it is difficult for the robot to determine which aspects of the task the person meant to correct during pHRI, and which features were changed unintentionally. 

\medskip

\noindent\textbf{Independent Variables.} We used a 2-by-2 factorial design and manipulated the \textit{learning strategy} with two levels (\emph{All-at-Once} and \emph{One-at-a-Time}), as well as the \textit{number of feature weights that need correction} (one feature weight and all the feature weights). Within the All-at-Once learning strategy the robot always updated all the feature weights after a single human interaction using the gradient update from Equation (\ref{eq:L5}). In the One-at-a-Time condition the robot chose the one feature that changed the most using Equation (\ref{eq:O3}), and then updated its feature weight according to Equation (\ref{eq:O5}). Both learning strategies leveraged Algorithm~\ref{alg:a1}, but with different update rules. By comparing these two versions of our approach we explore how robots should respond to noisy and imperfect human interactions.

\begin{figure}[t!]
\centering
\includegraphics[width=1\columnwidth]{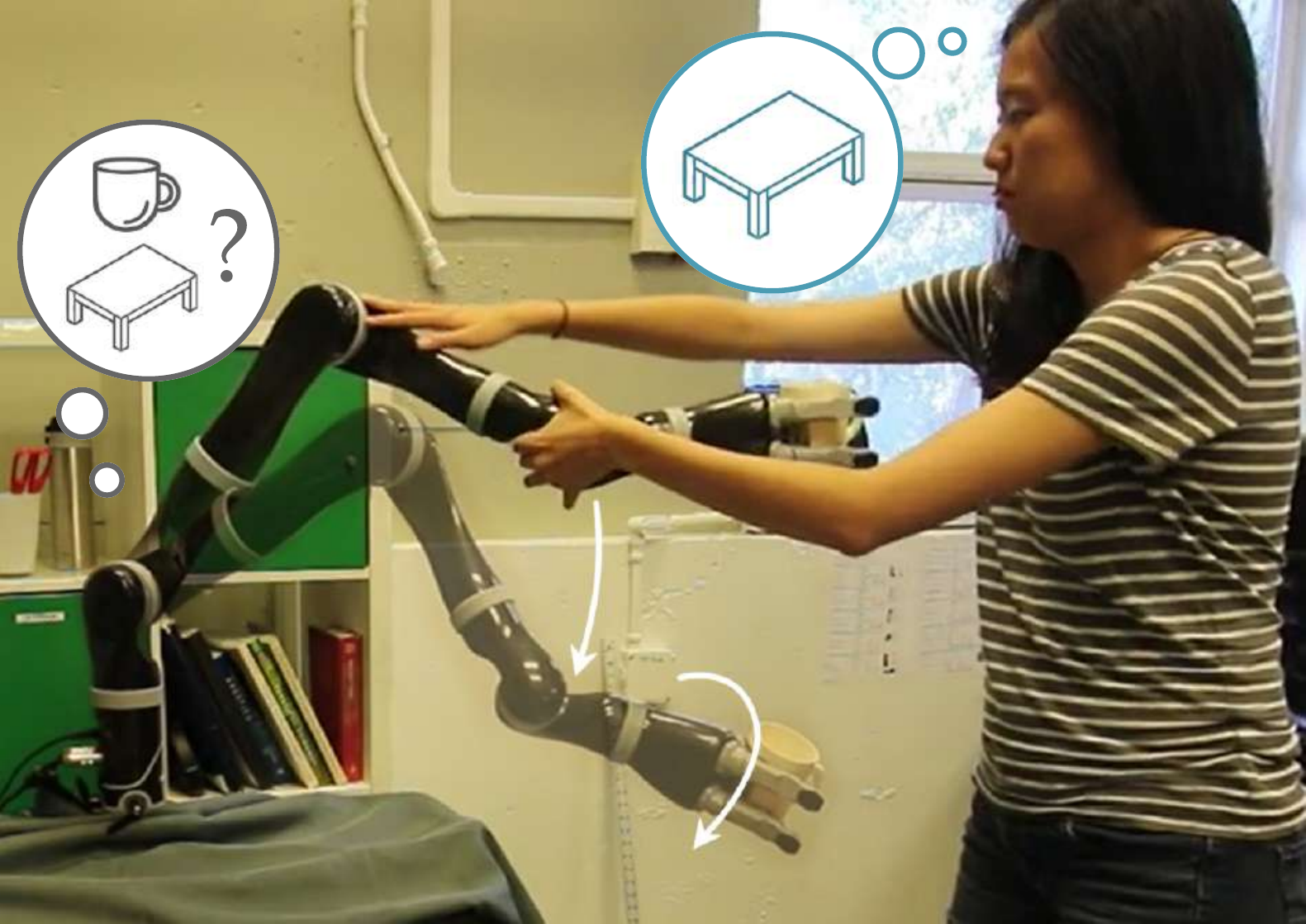}
\caption{When the robot has to trade-off between many task objectives, physically teaching the robot can be challenging for users. Here the human intends to teach the robot to move closer to the table. But the human's correction unintentionally tilts the angle of the cup: the robot must decide which parts of this correction to learn from and which parts to ignore.} \label{fig:hrifig}
\vspace{0.5em}
\end{figure}

\medskip

\noindent\textbf{Dependent Measures -- Objective.} Within this user study the robot carried a cup across a table. To analyze the objective performance of our two learning strategies, we split the objective measures into four categories:

\smallskip

\noindent\textit{Final Learned Reward:} These metrics measure how closely the learned reward matched the optimal reward by the end of the task (timestep $T$). We measured the dot product between the optimal and final reward vector:
$
\text{\emph{DotFinal}} = \theta \cdot \hat{\theta}^T
$.
We also analyzed the regret of the final learned reward, which is the weighted feature difference between the ideal trajectory and the learned trajectory:
$$\text{\emph{RegretFinal}} = \theta \cdot \Phi(\xi_{\theta})-\theta\cdot\Phi(\xi_{\hat{\theta}^T})$$
Lastly, we measured the individual feature differences (table and cup) between the ideal and final learned trajectories:
$$\text{\emph{TableDiffFinal}} = |\Phi_{table}(\xi_{\theta})-\Phi_{table}(\xi_{\hat{\theta}^T})|$$
$$\text{\emph{CupDiffFinal}} = |\Phi_{cup}(\xi_{\theta})-\Phi_{cup}(\xi_{\hat{\theta}^T})|$$

\smallskip

\noindent\textit{Learning Process:} Measures about the learning process, i.e., $\mathbf{\tilde{\theta}} = \{ \hat{\theta}^0,\hat{\theta}^1,\hdots,\hat{\theta}^T\}$, included the average dot product between the true reward and the estimated reward over time:
$$
\text{\emph{DotAvg}} = \frac{1}{T}\sum^T_{i=0}\theta \cdot \hat{\theta}^i
$$
We also measured the length of the $\tilde{\theta}$ path through weight space for both cup ($\tilde{\theta}_{cup}$) and table ($\tilde{\theta}_{table}$) weights. Finally, we computed the number of times the cup and table weights were updated in the opposite direction of the optimal $\theta$ (denoted by \textit{CupAway} and \textit{TableAway}).

\smallskip

\noindent\textit{Executed Trajectory:} For the actual trajectory that the robot executed, $\xi_{act}$, we measured the regret
$$\text{\emph{Regret}} =\theta \cdot \Phi(\xi_{\theta})-\theta\cdot \Phi(\xi_{act})$$ and the individual table and cup feature differences between the ideal and actual trajectory
$$\text{\emph{TableDiff}} = |\Phi_{table}(\xi_{\theta})-\Phi_{table}(\xi_{act})|$$
$$\text{\emph{CupDiff}} = |\Phi_{cup}(\xi_{\theta})-\Phi_{cup}(\xi_{act})|$$

\smallskip

\noindent\textit{Interaction:} Interaction measures on the forces applied by the human included the total interaction force, \emph{IactForce} = $\sum^T_{t=0}||u_h^t||_1$, and the total interaction time.

\medskip

\begin{figure*}[t!]
\begin{centering}
\includegraphics[width=1\textwidth]{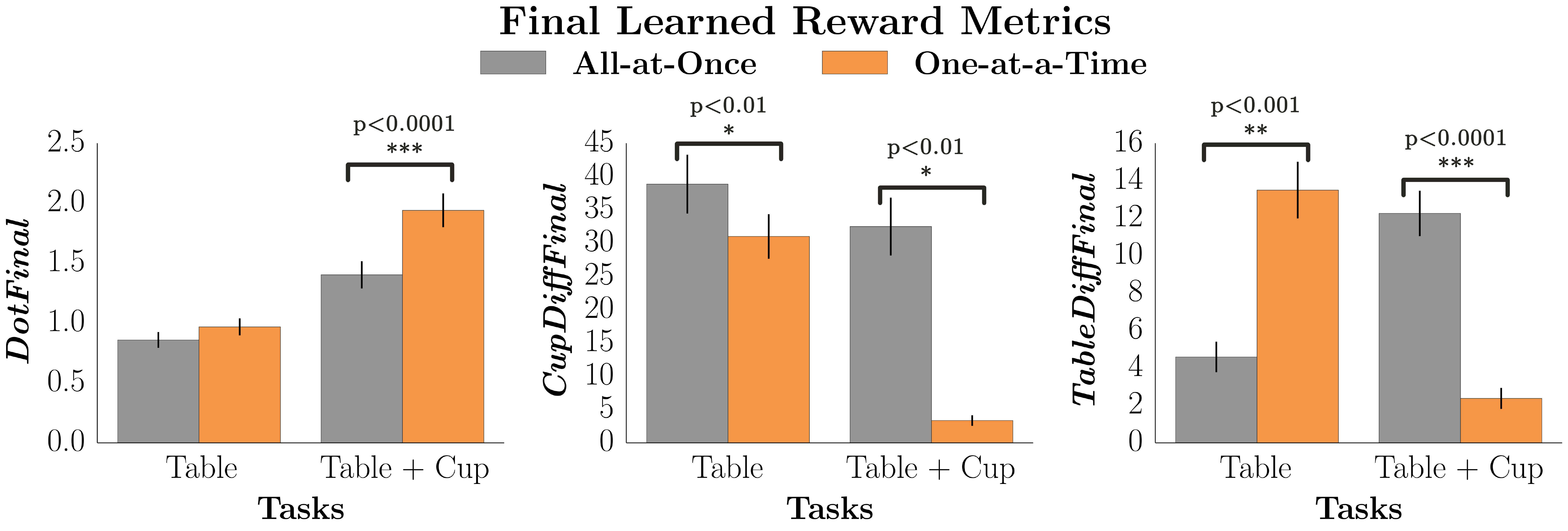}
\caption{How accurately the robot learned when using All-at-Once or One-at-a-Time. (Left) The final learned $\theta$ with One-at-a-Time is more aligned with the ideal $\theta$ on the Table+Cup task where the human had to correct multiple features. Looking at the individual feature errors: (Center) while the final cup feature was closer to ideal for One-at-a-Time on both tasks, (Right) All-at-Once learned a more accurate estimate of Table when the human only needed to teach a single feature. But we notice an interaction effect here: although One-at-a-Time got the Table wrong on the single feature task, it outperformed All-at-Once across the board when the human needed to adjust multiple features.}  \label{fig:dotDiff}
\end{centering}
\end{figure*}

\noindent\textbf{Dependent Measures -- Subjective.} After each of the four conditions we administered a 7-point Likert scale survey about the participant's interaction experience (see Table \ref{tab:likert_HRI} for the list of questions). We separated our survey items into four scales: success in teaching the robot about the task (succ), correctness of update (correct update), needing to undo corrections because the robot learned something wrong (undoing), and ease of undoing (undo ease).

\medskip

\noindent\textbf{Hypotheses.} 

\begin{displayquote}
\textbf{H3.} \emph{One-at-a-Time learning will increase the final learned reward, enable a better learning process, result in lower regret for the executed trajectory, and lead to less interaction effort and time as compared to All-at-Once.}
\end{displayquote}

\begin{displayquote}
\noindent\textbf{H4.} \emph{Participants will perceive the robot as more successful at accomplishing the task, better at learning, less likely to need undoing, and easier to correct if it did learn something wrong in the One-at-a-Time condition.}
\end{displayquote}

\medskip

\noindent\textbf{Tasks.} We designed two household manipulation tasks for the robot arm to perform within a shared workspace. A depiction of the these experimental tasks is shown in Fig.~\ref{fig:tasks_hri}. The robot's objective function consisted of three features: ``velocity," (the trajectory length), ``table'' (the distance from the table), and ``cup" (the orientation of the cup). We purposely selected features that were easy for participants to interpret so that they intuitively understood how to correct the robot. For each experimental task the robot carried a cup from a start pose to end pose with an \textit{initially incorrect objective}. Task 1 focused on participants having to correct a \textit{single aspect} of the objective, while Task 2 required them to correct \textit{all parts} of the objective.

In Task 1 the robot's objective had only \textit{one feature weight incorrect}. The robot's default trajectory took a cup from the participant and put it down on the table, but carried the cup too far above the table (see top of Fig.~\ref{fig:tasks_hri}). In Task 2 \textit{all the feature weights started out incorrect} in the robot's objective. The robot again took a cup from the participant and put it down on the table, but this time it initially grasped the cup at the wrong angle, and was also carrying the cup too high above the table (see bottom of Fig.~\ref{fig:tasks_hri}). 

\medskip

\noindent\textbf{Participants.} We used a within-subjects design and counterbalanced the order of the conditions during experiments. In total, twelve members of the UC Berkeley community ($4$ male, $7$ female, $1$ non-binary trans-masculine, age range $18$-$30$) provided informed written consent according to the approved IRB protocol before participating in this study. Eleven of the participants had technical backgrounds, and one did not. None of the participants had prior experience interacting with the robot used in our experiments. 

\medskip

\noindent\textbf{Procedure.} Before the start of the experiment participants performed a familiarization task to become more comfortable teaching the 7-DoF JACO2 robot with physical corrections. We here used the second task from our first experiment, where the robot carried a cup at an angle, and the human must correct the cup's orientation. During this familiarization task the robot's objective contained only one feature weight (cup). Afterwards, for each experimental task, the participants were shown the robot's initial trajectory as well as their desired trajectory. They were also told what aspects of the task the robot is aware of (cup orientation and distance to table), as well as which learning strategy they were interacting with (One-at-a-Time or All-at-Once). Participants were told the difference between the two learning strategies in order to minimize in-task learning effects. Importantly, we did \textit{not} tell participants to teach the robot in any specific way (like one aspect as a time); we only informed participants about how the robot reasons over their corrections.

\medskip

\noindent\textbf{Results -- Objective.} Here we summarize the results for each of our objective dependent measures.

\smallskip

\noindent\textit{Final Learned Reward.} We ran a factorial repeated-measures ANOVA with learning strategy and number of features as factors---and user ID as a random effect---for each of our objective metrics. Fig.~\ref{fig:dotDiff} summarizes our findings about the final learned weights $\hat{\theta}^T$ for both learning strategies.

For the final dot product with the true reward $\theta$, we found a significant main effect of the learning strategy ($F(1,81)=29.86$, $p<.0001$), but also an interaction effect with the number of features ($F(1,81)=13.07$, $p<.01$). The post-hoc analysis with Tukey HSD revealed that One-at-a-Time led to a higher dot product on Task 2 ($p<.0001$), but there was no significant difference on Task 1 (where One-at-a-Time led to slightly higher dot product).

We next looked at the final regret, i.e., the difference between the cost of the final learned trajectory and the cost of the ideal trajectory. For this metric we found an interaction effect, suggesting that One-at-a-Time led to lower regret for Task 2 but not for Task 1. Looking separately at the feature values for table and cup, we found that One-at-a-Time led to a significantly lower difference for the cup feature across the board ($F(1,81)=11.30$, $p<.01$, no interaction effect), but that One-at-a-Time only improved the difference for the table on Task 2 ($p<.0001$). Surprisingly, One-at-a-Time significantly increased the difference when the human only needed to correct a single feature ($p<.001$). 

Overall, we see that One-at-a-Time results in better final learning when the human needs to correct multiple features (Task 2). When the human only wants to correct a single feature (Task 1) the results are mixed: One-at-a-Time led to a significantly better result for the cup orientation, but a significantly worse result for the table distance.

\smallskip

\begin{figure}[t]
\begin{centering}
\subfigure[DotAvg measures the alignment between the learned $\hat{\theta}^t$ and the true $\theta$. In the task with only one wrong feature weight, there was no significant difference between the two methods in average dot product over time.  \label{fig:cupdiff}]{\includegraphics[width=.5\textwidth]{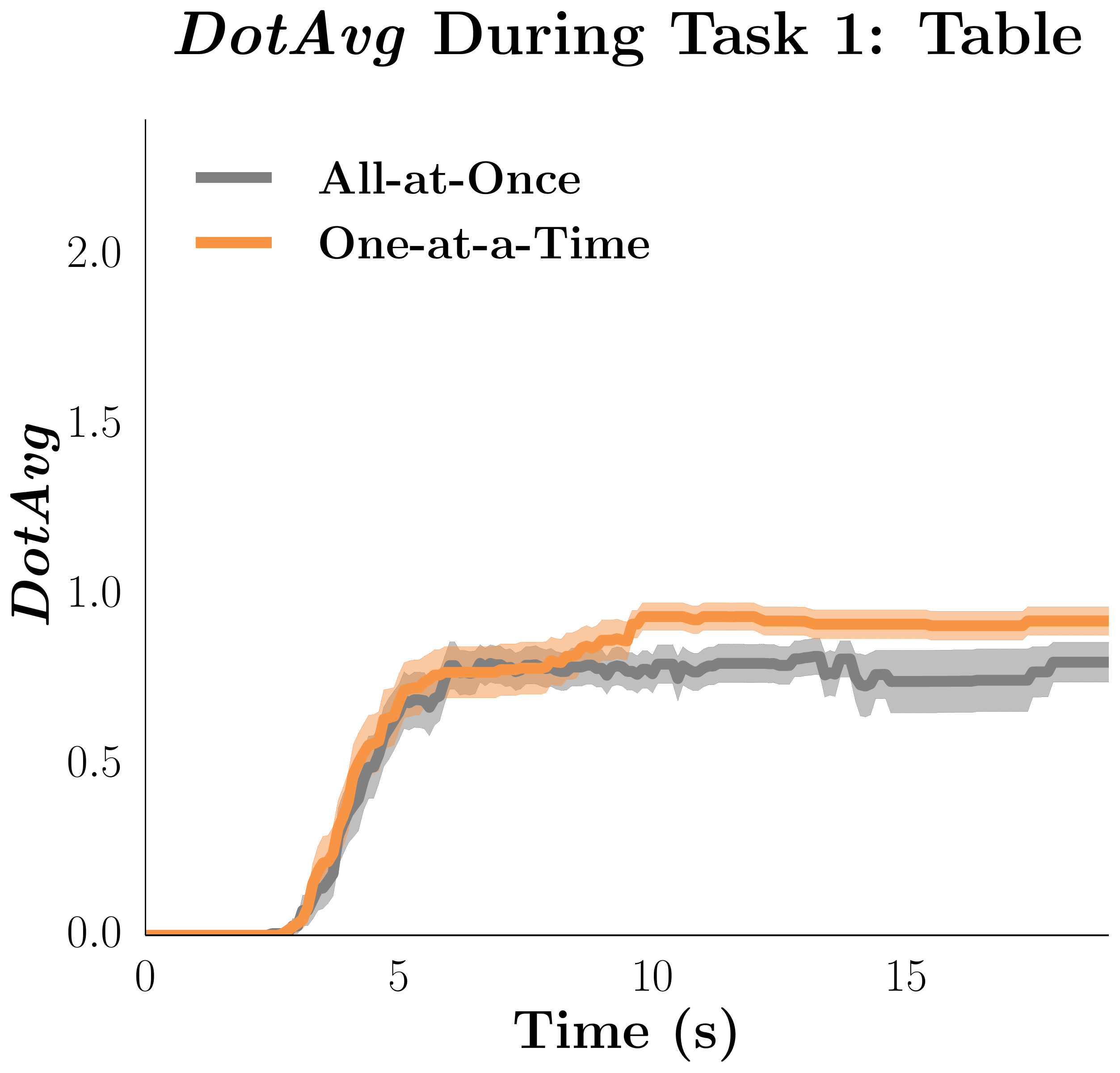}}\quad
\subfigure[In contrast to (a), when two feature weights are wrong One-at-a-Time outperformed All-at-Once. The dip in DotAvg for All-at-Once indicates that participants accidentally taught the robot the wrong thing and needed to undo their corrections. \label{fig:tablediff}]{\includegraphics[width=.5\textwidth]{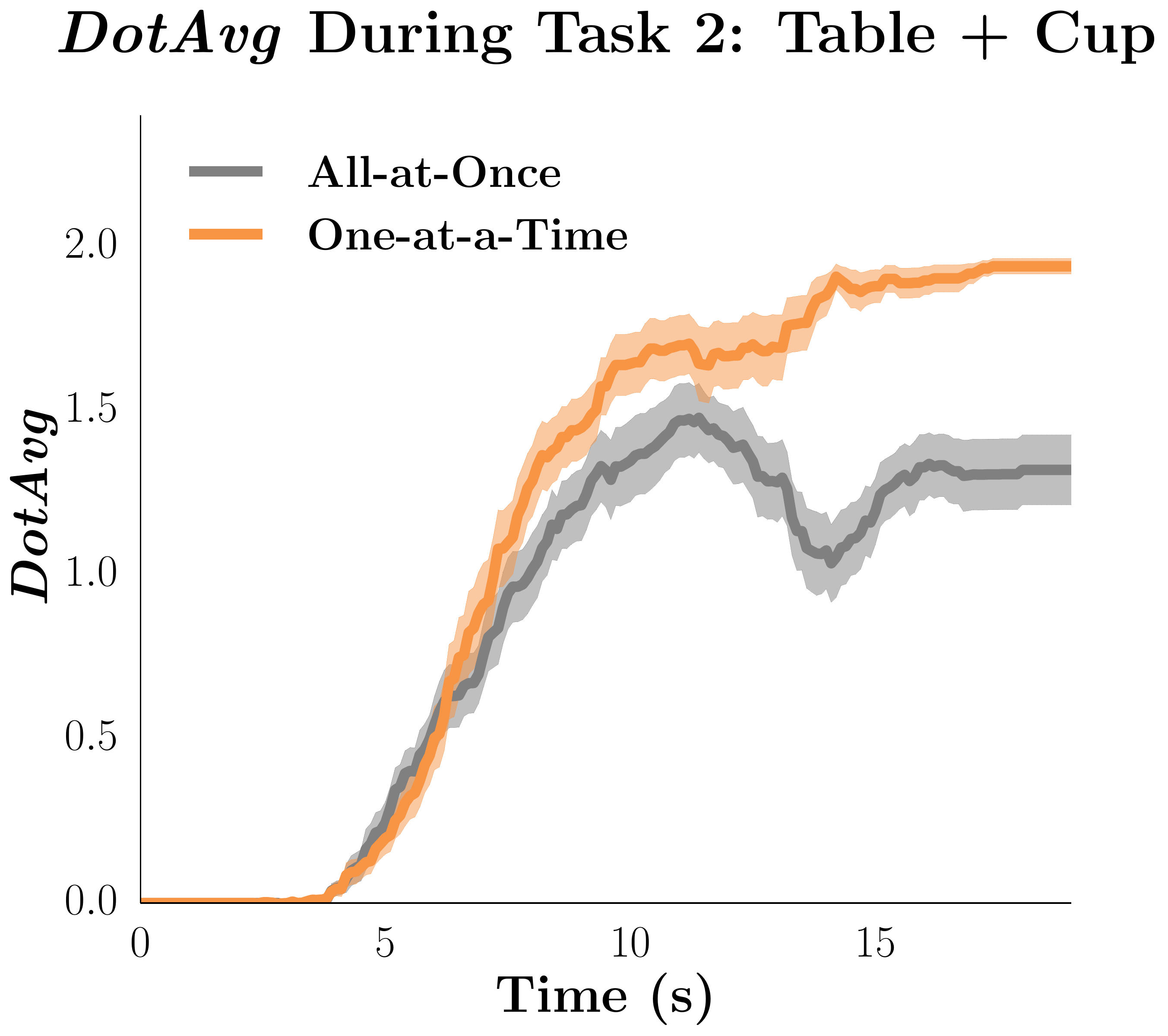}}
\caption{One-at-a-Time showed more consistent alignment between the learned objective, $\hat{\theta}^t$, and the ideal objective, $\theta$, when compared to All-at-Once. Contrasting (a) and (b), these results suggest that when the human needs to correct multiple aspects of the robot's behavior One-at-a-Time enables more accurate learning. We anticipate that most real-world tasks will require corrections of multiple features.} \label{fig:diffMetric}
\end{centering}
\end{figure}

\noindent\textit{Learning Process.} For the average dot product between the estimated and true reward over time, our analysis revealed almost identical outcomes as those reported for the final reward (see Fig.~\ref{fig:diffMetric}). Higher values of $DotAvg$ indicate the robot's estimate $\hat{\theta}$ is in the direction of the true parameters $\theta$. Differences in $DotAvg$ were negligible during Task 1, but One-at-a-Time outperformed All-at-Once during Task 2.

\begin{figure}[t!]
\includegraphics[width=0.5\textwidth]{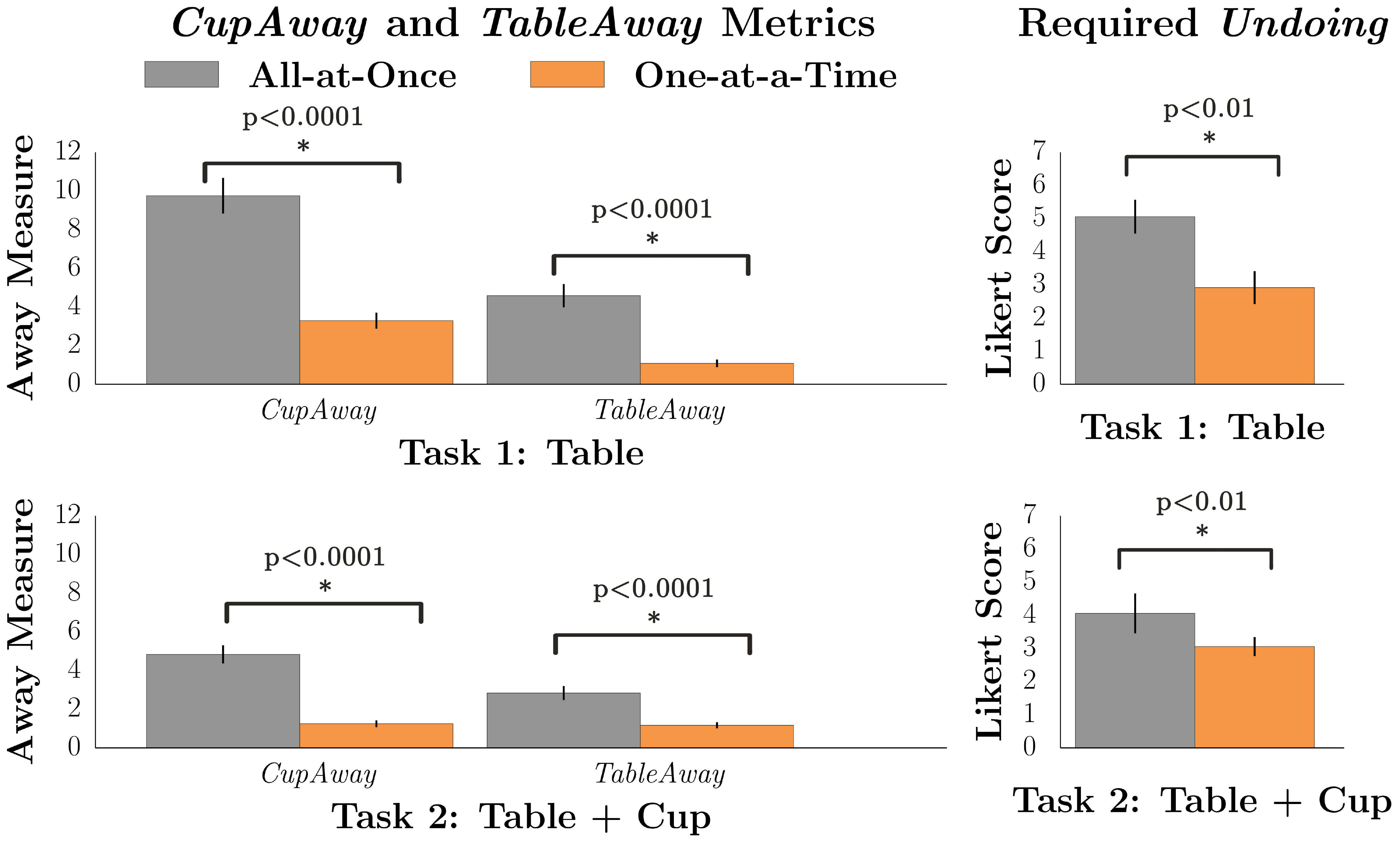}
\caption{How frequently participants made mistakes and had to undo their corrections. (Left) Humans working with One-at-a-Time made fewer corrections that caused the robot to learn the opposite of what they intended. This result was consistent across both tasks. (Right) These objective findings match our subjective Likert scale data. Participants thought the One-at-a-Time robot was less likely to learn the wrong thing and need an additional undoing action.} \label{fig:awayUndo}
\end{figure}

Next, we found that One-at-a-Time resulted in significantly fewer updates in the wrong direction for the cup weight ($F(1,81)=44.91$, $p<.0001$) and for the table weight ($F(1,81)=22.02$, $p<.0001$), with no interaction effect in either case. Fig.~\ref{fig:awayUndo} highlights these findings and their connection to the subjective user responses from Table~\ref{tab:likert_HRI} that are related to undoing.

Finally, looking at the length of the learned path $\tilde{\theta}$ through the space of feature weights, we found a main effect of learning strategy ($F(1,81)=26.82$, $p<.0001$), but also an interaction effect ($F(1,81)=6.55$, $p=.01$). The post-hoc analysis with Tukey HSD revealed that for Task 1 our One-at-a-Time approach resulted in a significantly shorter path through weight space ($p<.0001$). The path was also shorter during Task 2, but this difference was not significant. The effect was mainly due to the One-at-a-Time method resulting in a shorter path for the cup weight on Task 1, as revealed by the post-hoc analysis ($p<.0001$). 

Overall, we see that the quality of the learning process was significantly higher for the One-at-a-Time strategy across both tasks. When one aspect (Task 1) or all aspects (Task 2) of the objective were wrong, One-at-a-Time led to fewer weight updates in the wrong direction, and resulted in the learned reward over time being closer to the true reward.

\smallskip

\noindent\textbf{The Executed Trajectory.} We found no significant main effect of the learning strategy on the regret of the executed trajectory: the two strategies lead to relatively similar actual trajectories with respect to regret. Both regret as well as the feature differences from ideal for cup and table showed significant interaction effects.  
\smallskip

\noindent\textbf{Interaction Metrics.} We found no significant effects on interaction time or force.

\medskip

\noindent\textbf{Objective Results -- Summary.}
Taken together these results indicate that One-at-a-Time leads to a better overall learning process. On the more complex task where all the features must be corrected (Task 2), One-at-a-Time also leads to a better final learned reward. For the simpler task where only one feature must be corrected (Task 1), One-at-a-Time enables users to better avoid accidentally changing the initially correct weight (cup), but One-at-a-Time is not as good as the All-at-Once method at enabling users to properly correct the initially incorrect weight (table). Accordingly, our objective results partially support \textbf{H3}. Although updating one feature weight at a time does not improve task performance when only one aspect of the objective is wrong, reasoning about one feature weight at a time leads to significantly better learning and task performance when all aspects of the objective are wrong.

\medskip

\noindent\textbf{Results -- Subjective.} We ran a repeated measures ANOVA on the results of our participant survey. After testing the reliability of our four scales (see Table~\ref{tab:likert_HRI}), we found that the correct update and undoing scales were reliable, and so we grouped these into a combined score. The success (succ) scale had only a single question, and so grouping was not applicable here. Finally, we analyzed the two questions related to undoing ease (undo ease) individually because this specific scale was not reliable.

For the correct update scale we found a significant effect of learning strategy ($F(1,33)=5.09, p=0.031$), showing that participants perceived One-at-a-Time as better at updating the robot's objective according to their corrections. The undoing scale also showed a significant effect of learning strategy ($F(1,33)=10.35, p<0.01$), where One-at-a-Time was perceived as less likely to learn the wrong thing, which would then force the participants to undo their corrections. For both success and undoing ease scales we analyzed the questions Q1, Q9, and Q10 individually and found no significant effect of learning strategy.

\medskip

\noindent\textbf{Subjective Results -- Summary.}
The subjective data echoes some of our objective data results. Participants perceived that the robot with One-at-a-Time was better at correcting what they intended, and required less undoing due to unintended learning. We conclude that \textbf{H4} was partially supported.

{\renewcommand{\arraystretch}{1.5}
\begin{table}[t]
 \centering
 \adjustbox{max width=\textwidth}{
 \begin{tabular}{|c|l|c|}
 \hline
 \rowcolor{Gray}
 & \multicolumn{1}{c|}{\textbf{Questions}} & \multicolumn{1}{c|}{\textbf{Cronbach's $\alpha$}} \\
 \hline
    \parbox[t]{2mm}{\multirow{2}{*}{\rotatebox[origin=c]{90}{\textbf{succ}}}} & \multicolumn{1}{p{5cm}}{Q1: I successfully taught the robot how to do the task.} & \multirow{2}{*}{--} \\
    \hline
    \parbox[t]{2mm}{\multirow{8}{*}{\rotatebox[origin=c]{90}{\textbf{correct update}}}} & \multicolumn{1}{p{5cm}}{Q2: The robot correctly updated its understanding about aspects of the task that I did want to change.} & \multirow{8}{*}{.84} \\
    & \multicolumn{1}{p{5cm}}{Q3: The robot wrongly updated its understanding about aspects of the task I did NOT want to change.} &\\
    & \multicolumn{1}{p{5cm}}{Q4: The robot understood which aspects of the task I wanted to change, and how to change them.}  &\\
    & \multicolumn{1}{p{5cm}}{Q5: The robot misinterpreted my corrections.}  &\\
    \hline
    \parbox[t]{2mm}{\multirow{7}{*}{\rotatebox[origin=c]{90}{\textbf{undoing}}}} 
    & \multicolumn{1}{p{5cm}}{Q6: I had to try to undo corrections that I gave to the robot, because it learned the wrong thing.}  & \multirow{7}{*}{.93} \\
    & \multicolumn{1}{p{5cm}}{Q7: Sometimes my corrections were just meant to fix the effect of previous corrections I gave.}  & \\
    & \multicolumn{1}{p{5cm}}{Q8: I had to re-teach the robot about an aspect of the task that it started off knowing well.}  & \\
    \hline
    \parbox[t]{2mm}{\multirow{4}{*}{\rotatebox[origin=c]{90}{\textbf{undo ease}}}} 
    & \multicolumn{1}{p{5cm}}{Q9: When the robot learned something wrong, it was difficult for me to undo that.}   & \multirow{4}{*}{.66} \\
    & \multicolumn{1}{p{5cm}}{Q10: It was easy to re-correct the robot whenever it misunderstood a previous correction of mine.} &\\
    \hline
 \end{tabular}
 }
 \strut 
 \label{table}
 \caption{Likert scale questions from our user study comparing All-at-Once and One-at-a-Time. Questions were grouped into four categories: success in accomplishing the task (succ), whether the robot's update was what the human wanted (correct update), how often the human needing to undo corrections because of unintended learning (undoing), and how easy it was to undo a mistake (undo ease).} \label{tab:likert_HRI}
 \end{table}}
\section{Discussion}

In this work we recognize that when humans physically interact with and correct a robot's behavior their corrections become a source of information. This insight enables us to formulate pHRI as a partially observable dynamical system: the robot is unsure of its true objective function, and human interactions become observations about that latent objective. Solving this dynamical system results in robots that respond to pHRI in the \emph{optimal way}. These robots update their understanding of the task after each human interaction, and then change how they complete the rest of the current task based on this new understanding.

\medskip

\noindent\textbf{Approximations.} Directly applying our formalism to find the robot's optimal response to pHRI is generally not tractable in high-dimensional and continuous state and action spaces. We therefore derive an online approximation for robot learning and control. We first leverage the QMDP approximation \citep{littman1995learning} to separate the learning problem from the control problem, and then move from the policy level to the trajectory level. This results in two local optimization problems. In the first, the robot solves for an optimal trajectory given its MAP estimate of the task objective, and then tracks that trajectory using impedance control \citep{hogan1985impedance}. The second optimization problem occurs at timesteps when the human interacts: here the robot updates its estimate of the correct objective using online gradient descent \citep{bottou1998online}; this update rule is a special case of Coactive Learning \citep{jain2015learning,shivaswamy2015} and Maximum Margin Planning \citep{ratliff2006maximum}. Although we can practically think of the proposed algorithm as using impedance control to track a trajectory that is replanned after physical interactions, this approach ultimately derives from formulating pHRI as an instance of a POMDP.

Interestingly, this derivation enables us to interpret other state-of-the-art responses to pHRI as simplifications of our approximation. For example, if the robot never updates its estimate of the correct objective function (i.e., the robot never learns from pHRI), then our online approximation reduces to impedance control. Alternatively, if we treat the intended trajectory induced by the human's correction as the robot's trajectory (but do not update the robot's objective), then our approximation reduces to deforming the desired trajectory \citep{losey2018trajectory}. We compared our online approximation to both of these simplifications---impedance control and deformations---as well as to a more complete QMPD solution. During offline simulations we found that the performance loss between our learning method and the QMPD policy was negligible, but our method outperformed impedance control and trajectory deformations. During user studies with a 7-DoF robot, our learning approach resulted in decreased interaction time, effort, and cumulative trajectory cost when compared to an impedance controller. We also found that users believed the learning robot better understood their preferences, resulted in less interaction effort, and was more a collaborative partner than the impedance robot.

\medskip

\noindent\textbf{Unintended Corrections.} While we assert that the human's physical interactions are often intentional, we also recognize that physical interactions are inherently noisy and imperfect. When correcting a high DoF robot the human may adjust aspects of the robot's behavior that they did not intend to. If the robot treats every aspect of the human's correction as intentional this can result in unintended learning, which the human must then undo with additional corrections. In order to mitigate the effects of unintended corrections, and make the process of correcting robots through pHRI more intuitive for the end-user, we introduce a restriction to our online learning rule. More specifically, we assume that the robot should only learn about one aspect of the task from each human correction. During offline simulations we showed that this One-at-a-Time learning approach outperformed All-at-Once when the simulated user acted noisily: with All-at-Once, the noisy human unintentionally changed aspects of the robot's task which were already correct, but with Once-at-a-Time these unintended corrections were avoided.

Next, we performed a user study to compare our One-at-a-Time and All-at-Once learning strategies. Here the robot could reason over multiple features during two tasks: one task required correcting a single feature, and the other task required correcting multiple features of the robot’s objective. For the multiple feature task learning about one feature at a time was objectively superior: it led to a better final learning outcome, took a shorter path to the optimum, and had fewer incorrect inferences and human undoing along the way. But the results were not as clear for the single feature task: One-at-a-Time reduced unintended learning on the weights that were initially correct, but it hindered learning for the initially incorrect weights. Overall, study participants subjectively preferred One-at-a-Time to All-at-Once: they thought One-at-a-Time was better at learning the intended aspects of their corrections and required less undoing.

Based on these results, we hypothesize that the superior objective performance of One-at-a-Time was due to the increased complexity of the teaching task. It appears that only learning a single aspect at a time is more useful when the teaching task becomes more complex and requires that the human alter multiple parts of the robot's objective. When the teaching task is simple, however, and only requires one aspect of the objective to change, it is not yet clear whether One-at-a-Time is a better learning strategy.

\medskip

\noindent\textbf{Limitations.} Our work is a step towards understanding how robots should respond to pHRI. When selecting the approximations for online learning, as well as the method for inferring which feature to update in One-at-a-Time, we opt for approximations that are consistent to those in the existing literature. Future work and hardware advances may remove the need for some of the approximations we have leveraged.

Throughout our paper we assumed that the robot had access to the necessary task-related features. Moreover, during our user studies the robot's objective contained only two or three total features, and these features were intuitive to the human (e.g., ``distance-to-person"). In practice objective functions will have larger features sets and may include task-related features that are non-intuitive to the human: additional work is needed to investigate how well our learning strategies perform in these cases.

Finally, solutions that can handle dynamical aspects---like preferences about the timing of the robot's trajectory---would require a different approach for inferring the intended human trajectory. Here it may actually be necessary to return from the trajectory space to the policy space.
\section{Conclusion}

In this work we present an online, in-task response to pHRI that treats human interactions as intentional. We first formulate the problem of responding to pHRI as a partially observable dynamical system, where solving this system defines the \emph{optimal way} for the robot to react. Unfortunately, this formalism is not directly applicable because we require online solutions in high-dimensional and continuous state, action, and belief spaces. We therefore derive an approximate solution for real-time learning and control. During offline simulations we compared our approximate learning method to a complete solution and state-of-the-art baselines, which are actually simplifications of our approach. We perform two separate user studies on a 7-DoF robot arm to determine (a) whether learning from pHRI is useful and (b) how the robot should learn from physical human interactions. While these simulations and user studies indicate the benefits of our approach, we recognize that this work is only a first step towards leveraging the implicit communication present during human-robot interactions.

\bibliographystyle{SageH}
\bibliography{citations}

\end{document}